\documentclass[10pt,twocolumn,letterpaper]{article}

\usepackage[pagenumbers]{wacv} % To force page numbers, e.g. for an arXiv version

\usepackage{graphicx}
\usepackage{amsmath}
\usepackage{amssymb}
\usepackage{booktabs}

\usepackage{multirow}
\usepackage{multicol}
\usepackage{dblfloatfix}
\usepackage{algpseudocode}
\usepackage{algorithm}
\usepackage{enumitem}
\usepackage{lipsum} % for dummy text
\usepackage{wrapfig} % for wrapping text around figures
\usepackage[dvipsnames]{xcolor}
\usepackage{color, colortbl}
\usepackage{makecell}
\usepackage{pifont}
\usepackage{amsmath}

\newcommand{\AlgoName}{CODIP }
\newcommand{\AlgoNameNoSpace}{CODIP}
\newcommand{\AlgoNameTop}{$\text{CODIP}_{\text{Top-}k}$ }
\newcommand{\AlgoNameTopNoSpace}{$\text{CODIP}_{\text{Top-}k}$}
\definecolor{gray}{RGB}{0, 0, 0}
\definecolor{golden}{RGB}{255, 215, 0}
\usepackage[pagebackref,breaklinks,colorlinks]{hyperref}

\usepackage[capitalize]{cleveref}
\crefname{section}{Sec.}{Secs.}
\Crefname{section}{Section}{Sections}
\Crefname{table}{Table}{Tables}
\crefname{table}{Tab.}{Tabs.}

\def\wacvPaperID{282} % *** Enter the WACV Paper ID here
\def\confName{WACV}
\def\confYear{2025}

\begin{document}

\title{Class-Conditioned Transformation for Enhanced Robust Image Classification}

\author{
    Tsachi Blau\textsuperscript{\dag}\thanks{\texttt{tsachiblau@campus.technion.ac.il}} \quad 
    Roy Ganz\textsuperscript{\dag} \quad 
    Chaim Baskin\textsuperscript{\ddag} \quad 
    Michael Elad\textsuperscript{\dag} \quad 
    Alex M. Bronstein\textsuperscript{\dag}
}

% \and
\maketitle

\renewcommand{\thefootnote}{\fnsymbol{footnote}}
\footnotetext[2]{Department of Computer Science Technion - Israel Institute of Technology}
\footnotetext[3]{School of Electrical and Computer Engineering - Ben-Gurion University of the Negev}

%%%%%%%%% ABSTRACT
\begin{abstract}
    Robust classification methods predominantly concentrate on algorithms that address a specific threat model, resulting in ineffective defenses against other threat models. Real-world applications are exposed to this vulnerability, as malicious attackers might exploit alternative threat models. In this work, we propose a novel test-time threat model agnostic algorithm that enhances Adversarial-Trained (AT) models. Our method operates through COnditional image transformation and DIstance-based Prediction (CODIP) and includes two main steps: First, we transform the input image into each dataset class, where the input image might be either clean or attacked. Next, we make a prediction based on the shortest transformed distance. The conditional transformation utilizes the perceptually aligned gradients property possessed by AT models and, as a result, eliminates the need for additional models or additional training. Moreover, it allows users to choose the desired balance between clean and robust accuracy without training. The proposed method achieves state-of-the-art results demonstrated through extensive experiments on various models, AT methods, datasets, and attack types. Notably, applying CODIP leads to substantial robust accuracy improvement of up to $+23\%$, $+20\%$, $+26\%$, and $+22\%$ on CIFAR10, CIFAR100, ImageNet and Flowers datasets, respectively.   
    \\For more details, visit the \href{https://github.com/tsachiblau/Class-Conditioned-Transformation-for-Enhanced-Robust-Image-Classification}{project page}.
\end{abstract}

\begin{figure*}[t!]
    \centering
    \includegraphics[width=1.0\textwidth]{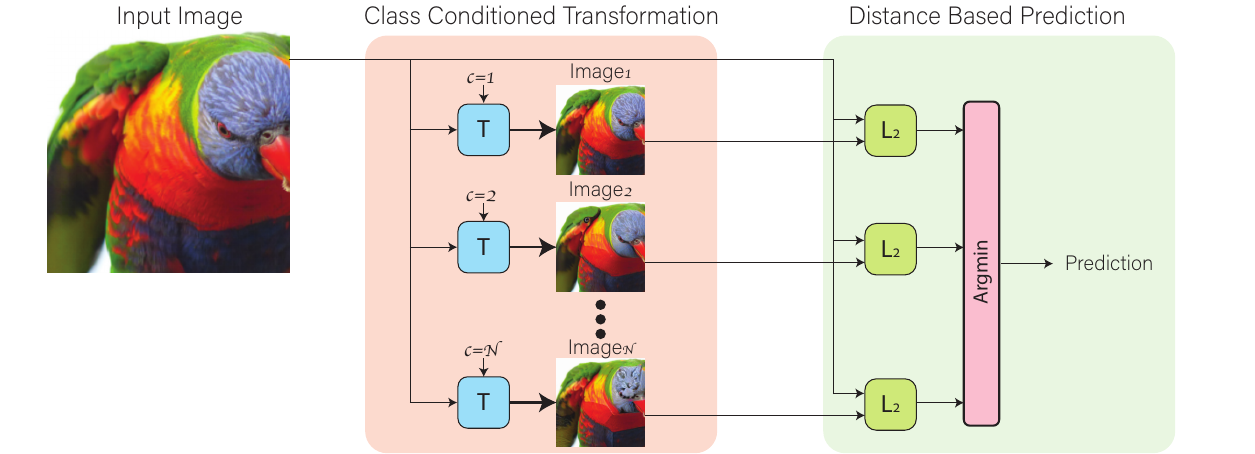}
    \caption{
        \textbf{An Overview of \AlgoName}
            At first, the input image (clean or attacked) is class conditioned transformed through $\{\text{T}(\cdot|1), \dots,\text{T}(\cdot|N)\}$ to each one of the dataset classes, creating images $\{\text{Image}_1, \dots, \text{Image}_N\}$. Next, the $\ell_2$ distance is calculated between the input image and the transformed images $\{\text{Image}_1, \dots, \text{Image}_N\}$, and prediction is made based on the shortest distance.
    }
    \label{fig:teaser}
\end{figure*}

\section{Introduction}
\label{sec:intro}

% explain AT
Adversarial attacks are a significant concern in the field of computer vision, where maliciously designed perturbations are injected into an image to fool classifiers \cite{carlini2017adversarial,goodfellow2014explaining,kurakin2016adversarial, athalye2018synthesizing,biggio2013evasion,szegedy2013intriguing,hosseini2017google,dodge2017study,geirhos2017comparing,temel2018cure,temel2018traffic, kurakin2018adversarial,nguyen2015deep}. 
One common method of generating an adversarial example is through an iterative optimization process that searches for a norm-bounded perturbation $||\delta||_{p} \leq \epsilon$ that is added to the input image. 
The properties of the perturbation are determined by a threat model, characterized by the choice of the radius $\epsilon$ and the norm $\ell_p$ (typical choices for the latter are $p\in\{1,2,\infty\})$.
Following the discovery of adversarial attacks several defenses were introduced, such as Adversarial Training (AT) \cite{gowal2020uncovering,rebuffi2021fixing,madry2017towards,carlini2017adversarial,zhang2019theoretically,goodfellow2014explaining,kannan2018adversarial}.
AT incorporates adversarial examples into the training process alongside the true labels, training the model to correctly classify malicious examples.

%test time methods 
Test-time methods \cite{cohen2019certified, raff2019barrage, perez2021enhancing, schwinn2022improving, wu2021attacking} aim to improve trained AT classifiers by changing their inference methodology. 
These methods change the standard prediction process where an instance is fed into a model that outputs a class prediction. Instead, they follow an inference algorithm that strengthens the prediction at the cost of additional inference time. While these methods are appealing, they have been used only to enhance the performance of a model using the same threat model used during training.
Test-time methods have yet to be applied for the threat model-agnostic problem \cite{laidlaw2020perceptual, maini2020adversarial, kaufmann2019testing, blau2022threat}, methods that defend a model against threat models that were not used during training.
Threat model-agnostic property is desirable in real-world applications as the attacker is not limited to a single type of attack and will likely exploit the model's weaknesses, using threat models that were not used for training.

% motivation
One property of adversarial attacks is that they do not substantially alter the input image $x$, as the perturbation $\delta$ is usually imperceptible.
With this motivation in mind, we hypothesize that if we transform the attacked image towards the true class, the transformed image remains close to the input image, while, contrarily, a class-conditioned transformation toward a different class would lead to significant semantic changes to the input image.
Hence, we conceive of a solution that operates through an optimal class-conditioned transformation with a regularization, which keeps the transformed image close to the input image.
Unfortunately, such an optimal transformation does not exist and transformations usually require an additional model.

We show that such a transformation can be performed using the recently discovered perceptually aligned gradients (PAG) property possessed by AT models \cite{engstrom2019adversarial, etmann2019connection,ross2018improving,tsipras2018robustness}. 
Simply put, the gradients of a classifier with respect to the input image manifest an intelligible spatial structure perceptually similar to that of some class of the dataset. 
As a result, maximizing the conditional probability of such classifiers towards some target class, via a gradient-based method, results in class-related semantic visual features.

% our method overview
In this work, we propose COnditional transformation and DIstance-based Prediction (\AlgoNameNoSpace), a novel test-time threat model-agnostic defense that operates through two steps, as depicted in \cref{fig:teaser}.
First, we transform the input image into each one of the dataset classes, while regularizing the transformation, creating $\{\text{Image}_1 , \dots, \text{Image}_N\}$.
The transformation is regularized since we want to change the class of the input image while remaining as close as possible to the input image.
Moreover, the transformation utilizes the PAG property possessed by any AT model and spares the need for additional models or training. 
Next, we classify based on the shortest distance between the input image and the transformed images $\{\text{Image}_1 , \dots, \text{Image}_N\}$ with the assumption that the input image needs to cross a shorter distance to turn into the true class.

% motivation exp + key contributions
Our method can be applied to any differentiable classifier. Furthermore, it also equips the user with control over the clean-robust accuracy trade-off and enhances the model's performance for both seen and unseen threat models.
Additionally, we propose \AlgoNameTop, an efficient algorithm that filters the lowest probability classes, leading to speed up of inference time.  
We validate our method through extensive experiments on various AT methods, classifier architectures, and datasets. 
We empirically demonstrate that \AlgoName boosts the accuracy of any differentiable AT classifier, convolution or transformer based, across a variety of attacks, white-box or black-box, and generalizes well to unseen ones. 
Specifically, applying \AlgoName leads to substantial improvement of up to $+23\%$, $+20\%$, $+26\%$, and $+22\%$ on CIFAR10, CIFAR100, ImageNet and Flowers datasets, leading to state-of-the-art performance.

To summarize, the key contributions of our work are:
\begin{itemize}[nolistsep,leftmargin=*]
    \item \AlgoName, a novel test-time algorithm that operates through conditional transformation and distance-based prediction, utilizing the PAG property and does not require an additional transformation model or training.  
    \item  \AlgoNameTop, an efficient algorithm that speeds up the inference time.
    \item A controllable clean-robust accuracy trade-off mechanism, that can be adapted by the user as needed, without training.
    \item State-of-the-art results for seen and unseen threat models, white-box or black-box. Demonstrated over multiple models, convolution or transformer based, AT methods, datasets, and provide extensive ablations to each component of our method.
\end{itemize}

% \begin{figure*}[ht!]
%     % \vspace{-100pt}
%     \centering
%     \includegraphics[width=1.\textwidth]{images/baselines_comparison.pdf}
%     \caption{
%     \textbf{Baselines Comparison} Performance comparison between \AlgoName and other test-time methods.
%     The plot presents the average accuracy of four different attacks.
%     The compared methods are Base (the AT classifier), TTE \cite{perez2021enhancing}, DRQ \cite{schwinn2022improving}, and \AlgoNameNoSpace.
%     The AT methods are \cite{madry2017towards}, Rebuffi \emph{et al.} \cite{rebuffi2021fixing} and Gowal \emph{et al.} \cite{gowal2020uncovering}, which were trained with the threat model specified below each method.
%     }
%     \label{fig:baselines_comparison}
% \end{figure*}

\section{Related work}
\label{sec:related}

\paragraph{\normalfont\textbf{Adversarial Training}} methods \cite{madry2017towards,carlini2017adversarial,goodfellow2014explaining} were first introduced to robustify a classifier against a specific threat model. 
However, a group of works \cite{kaufmann2019testing,brown2018unrestricted} have shown that such adversarially trained classifiers are less effective against unseen attacks. 
This finding has led to research that focuses on defenses for unseen attacks by training a model on a group of threat models.
The work done by Jordan \emph{et al.} \cite{jordan2019quantifying} suggests an attack that targets both texture and geometry, while Maini \emph{et al.} \cite{maini2020adversarial} suggests a combination of PGD-based attacks.
Another attempt was made by Laidlaw \emph{et al.} \cite{laidlaw2020perceptual}, which provides a latent space norm-bounded AT.
This line of work, however, has two major drawbacks.
First, it leads to inferior performance for every threat model compared to training a designated model for each specific threat model. 
Second, even a group of threat models cannot cover all possible threats, leaving the model vulnerable to attacks not included in the training.

\paragraph{\normalfont\textbf{Generative Model-Based Defences}} are methods that aim at cleaning the attack from the image, shifting the image back to the image manifold, before feeding it into the classifier for classification. 
The work by Song \emph{et al.} \cite{song2017pixeldefend} purifies the image using pixelCNN and \cite{du2019implicit} suggest restoring the corrupt image with EBM. 
Hill \emph{et al.} ~\cite{hill2020stochastic} suggest to apply long-run Langevin sampling and Yoon \emph{et al.} ~\cite{yoon2021adversarial} offer gradient ascent score based-model. 
The work done by Blau \emph{et al.} \cite{blau2022threat} combines the two worlds of generative model and threat model agnostic defense, using a vanilla-trained diffusion model and classifier.
While these methods are close in spirit to ours in the sense that they perform a transformation over the attacked image, they all require additional models to perform the transformation.

\paragraph{\normalfont\textbf{Test-Time Methods}} improve the performance of trained classifiers solely by using an inference phase of the neural network, while not requiring specialized training. 
However, they may result in longer inference times. 
Cohen \emph{et al.} \cite{cohen2019certified} and Raff \emph{et al.} \cite{raff2019barrage} suggest multiple realizations of the augmented image to be performed during test-time.
Cohen \emph{et al.} \cite{cohen2019certified} suggest augmenting with Gaussian noise, while Raff \emph{et al.} \cite{raff2019barrage} advocate using other types of augmentation; both methods involve averaging the classification predictions from the augmented images. 
Perez \emph{et al.} \cite{perez2021enhancing} perform test augmentation and predict the average. 
Schwinn \emph{et al.} \cite{schwinn2022improving} suggest analyzing the robustness of a local decision region near the attacked image, and Wu \emph{et al.} \cite{wu2021attacking} suggest to attack towards every class. 
The methods proposed by \cite{cohen2019certified,raff2019barrage} require fine-tuning the model, while \cite{schwinn2022improving,wu2021attacking} require neither fine-tuning nor access to the training data.

There are several distinctions between our approach and existing test-time methods. 
Firstly, while these methods focus on improving the performance of the threat model used for training the AT model, they fail to address the crucial issue of threat model-agnostic attacks. 
Secondly, unlike these methods, which rely on the classifier to make predictions, our predictions are based on measuring the distance between the input image and all the transformed images. 
Lastly, the works of \cite{schwinn2022improving,wu2021attacking}, which are most similar to our method, operate only in a local neighborhood around the input image, preventing the model from making significant semantic changes to the image. 
In contrast, our model is not limited to a specific neighborhood around the input image, as it transforms the image and alters its semantic appearance.

\section{Our method}
\label{sec:our_method}
In this section, we supply an overview of our method.
Next, we offer a detailed explanation that goes through the motivation and each part of our algorithm. 
Finally, we present an efficient alternative for our method.

\begin{figure*}[th]
    \centering
    \includegraphics[width=\textwidth]{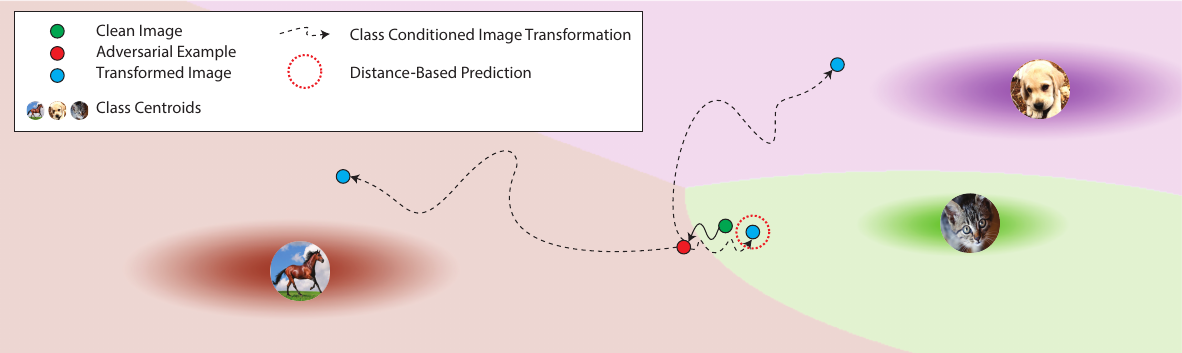}
    \caption{
      \textbf{A Comparison Between the Classifier and \AlgoName Decision Rules}
    The background color of the image describes the classifier's classification rules, and the intensity describes the classifier's certainty.
    The clean image (green dot) is attacked (red dot), leading to a wrong classification.
    In contrast, \AlgoName predicts based on the shortest transformation.
    It operates in two steps: First, it class conditioned transform (dotted arrow) the attacked image towards each one of the datasets's classes (blue dots). 
    Next, prediction is made based on the shortest distance between the attacked image and the transformed images, highlighted by a red dotted circle.
    }
    \label{fig:method}
\end{figure*}

\subsection{Method Overview}
We propose a test-time classification method that enhances an AT classification model for seen and unseen attacks.
Our method does not operate in the standard classification methodology $f : x \rightarrow y$, where $f(\cdot)$ is the classifier, $x\in R^{H\text{x}W\text{x}3}$ is the input image, $y \in [0,1]^N$ is the prediction probability vector, and $N$ is the number of classes of the dataset.

Instead, our method operates through two phases, as illustrated in \cref{fig:method}: 
First, we get an input image (which can be either clean or attacked) and transform it $N$ times, each time to a different class of the dataset.
The transformation utilizes the PAG property of the AT classifier and, hence, does not require additional architecture or additional training. 
The transformation is performed through an iterative class-conditioned regularized optimization process.
For every class, we perform $M$ such optimization steps, in each step we follow two objectives.
First, maximizing the classifier probability towards the current class.
Second, regularizing the transformation so the transformed image will stay close to the input image $x$.

\begin{algorithm*}[t!]
    \caption{CODIP}
    \label{alg:CODIP}
    \hspace*{\algorithmicindent}\textbf{Input} \text{   }classifier $f(\cdot)$, input image $x$, step size $\alpha$, regularization coefficient $\gamma$, \\ \hspace*{\algorithmicindent}\hspace*{\algorithmicindent} \hspace*{\algorithmicindent} number of iterations $M$, number of dataset's classes $N$ 
    \begin{algorithmic}[1]
    \Procedure{CODIP}{}
        \For{\texttt{$i \in  C = \{1,\dots, N\}$}} \Comment{Iterate over \# classes} \label{alg:loop_over_classes}
            \State $\mathbf{T}_{i 0} \gets 0$  \Comment{Initialize the $i^{th}$ transformation} \label{alg:CODIP_init}
            \For{\texttt{$j$ in $0:M-1$}} \Comment{Iterate over \# steps}
                \State $\mathbf{G}_{ij} \gets \nabla_{\mathbf{T}_{ij}}\left[ L_{CE}\left( f \left( x+ \mathbf{T}_{ij} \right), i \right) + \gamma \, \| \mathbf{T}_{ij} \|_2^2 \right]$ \label{alg:CODIP_calc_grad}
                \State $\mathbf{T}_{ij+1} \gets \Pi \left( \mathbf{T}_{ij} - \alpha \, \frac{\mathbf{G}_{ij}}{\| \mathbf{G}_{ij} \|_2 } \right)$ \label{alg:CODIP_update_trans}
            \EndFor
        \EndFor
    \State $\mathbf{d} \gets \|\mathbf{T}_{1:N M}\|_{2}$ \Comment{Update the transformation distances for each class} \label{alg:CODIP_clac_distance}
    \State $\hat{y} = \mathrm{arg}\min_{i=1,\dots,N} \mathbf{d}_i$ \Comment{Classify based on the shortest distance} \label{alg:CODIP_prediction}
    \State \Return $\hat{y}$
    \EndProcedure
    
    \end{algorithmic}
    The operator $\Pi(z)$ projects $z$ onto the image domain $\mathbb{R}^{H \times W \times 3} \in [0,1]^{H  \times W \times 3}$ by clipping its values. 
\end{algorithm*}

\subsection{\AlgoName}
In what follows, we detail our method summarized in \cref{alg:CODIP}:
We perform $N$ transformations towards each one of the dataset's classes, and we initialize the $i^{th}$ class transformation to zero $\mathbf{T}_{i0}$, in \cref{alg:CODIP_init}. 
Next, we start the transformation, which operates through $M$ gradient descent steps. 
In each iteration, we calculate the gradient of our objective and store it in $\mathbf{G}_{ij}$ in \cref{alg:CODIP_calc_grad}. 
Next, in \cref{alg:CODIP_update_trans}, we perform a gradient descent step in the direction of $\mathbf{G}_{ij}$, the step of size $\alpha$ which is a hyper-parameter. 
It is essential to normalize the gradient step since we want to make an even progress in each step, regardless of the gradient size.
Without normalization, the step size might be very small, which will prevent us from making progress during the transformation.
Before finishing the update step, in \cref{alg:CODIP_update_trans}, we perform a projection back to the image domain to $R^{H \times W \times 3} \in [0,1]^{H \times W \times 3}$. 
When finishing the transformation's iterations, we calculate the transformation distance using $\ell_2$ over the $M^{th}$ column of the transformation matrix, and we store it in the distance vector $\mathbf{d}$, in \cref{alg:CODIP_clac_distance}. 
Finally, when the process is over we predict according to the shortest transformation.

\begin{table*}[hb!]
\begin{center}
\vspace{0pt}
\caption{
    \textbf{CIFAR10 and CIFAR100 Results} The performance of test-time methods over two datasets: CIFAR10 and CIFAR100.  
    The results are grouped per AT model, evaluating different test-time methods: `Base' in which we do not use any test-time method, followed by other test-time methods.
    % \vspace{-30pt}
}
\resizebox{\textwidth}{!}{%

\begin{tabular}{lllllcccccc}
\hline\noalign{\smallskip}\hline

\rowcolor{gray!5}  &  &  &  &  &  & \multicolumn{4}{c}{Attack}\\
\rowcolor{gray!5} & & & & & & \multicolumn{2}{c}{$L_{\infty}$} & \multicolumn{2}{c}{$L_{2}$}\\
\rowcolor{gray!5} \multirow{-3}{*}{Dataset} & \multirow{-3}{*}{AT Method} & \multirow{-3}{*}{\makecell{Trained Threat \\ Model}} & \multirow{-3}{*}{Architecture}  & \multirow{-3}{*}{\makecell{Test-Time \\ Method}}  & \multirow{-3}{*}{Clean} & $8/255$ & $16/255$ & $0.5$ & $1.0$\\

\hline\noalign{\smallskip}\hline\noalign{\smallskip}

\multirow{17}{*}{CIFAR10} & PAT \cite{laidlaw2020perceptual} & & RN50 & & $71.60\%$ & $28.70\%$ & $-$ & $33.30\%$ & $-$\\

\cline{2-10}\noalign{\smallskip}

& \multirow{5}{*}{AT \cite{madry2017towards}} & \multirow{5}{*}{$L_{2}, \epsilon=0.5$} & \multirow{5}{*}{RN50} & Base & $90.83\%$ & $29.04\%$ & $00.93\%$ & $69.24\%$ & $36.21\%$\\

& & & &  RSmooth \cite{cohen2019certified} & $89.43\%$ &  $30.84\%$ & $01.23\%$ & $68.94\%$ & $38.53\%$\\

& & & &  TTE \cite{perez2021enhancing} & $\textbf{90.99}\%$ &  $36.41\%$ & $02.40\%$ & $71.90\%$ & $41.18\%$\\

& & & & DRQ \cite{schwinn2022improving} & $88.79\%$ & $45.37\%$ & $07.09\%$ & $77.56\%$ & $51.28\%$\\

& & & & \cellcolor{golden!10} \AlgoName  & \cellcolor{golden!10} $87.40\%$ & \cellcolor{golden!10} $\textbf{51.66\%}$ & \cellcolor{golden!10} $\textbf{14.96\%}$ & \cellcolor{golden!10} $\textbf{78.66\%}$ & \cellcolor{golden!10} $\textbf{59.82\%}$\\

\cline{2-10}\noalign{\smallskip}

& \multirow{4}{*}{Rebuffi \emph{et al.} \cite{rebuffi2021fixing}} & \multirow{4}{*}{$L_{2}, \epsilon=0.5$} & \multirow{4}{*}{WRN28-10} & Base & $\textbf{91.79\%}$ & $47.85\%$ & $05.00\%$ & $78.80\%$ & $54.73\%$\\

& & & &  TTE \cite{perez2021enhancing} & $91.59\%$ & $50.49\%$ & $06.79\%$ & $79.18\%$ & $55.38\%$\\

& & & &  DRQ \cite{schwinn2022improving} & $90.99\%$ & $58.66\%$ & $\textbf{13.69\%}$ & $84.12\%$ & $64.69\%$\\

& & & &  \cellcolor{golden!10} \AlgoName & \cellcolor{golden!10} $88.23\%$ & \cellcolor{golden!10} $\textbf{59.99\%}$ & \cellcolor{golden!10} $11.45\%$ & \cellcolor{golden!10} $\textbf{85.56\%}$ & \cellcolor{golden!10} $\textbf{66.15\%}$\\

\cline{2-10}\noalign{\smallskip}

& \multirow{4}{*}{Rebuffi \emph{et al.} \cite{rebuffi2021fixing}} & \multirow{4}{*}{$L_{\infty}, \epsilon=8/255$} & \multirow{4}{*}{WRN28-10} & Base & $\textbf{87.33\%}$ & $60.77\%$ & $25.44\%$ & $66.72\%$ & $35.01\%$\\

 & & & & TTE \cite{perez2021enhancing} & $87.30\%$ & $61.52\%$ & $27.50\%$ & $66.88\%$ & $36.07\%$\\

& & & & DRQ \cite{schwinn2022improving} & $87.17\%$ & $66.23\%$ & $33.62\%$ & $72.24\%$ & $44.56\%$\\

& & & & \cellcolor{golden!10} \AlgoName & \cellcolor{golden!10} $85.00\%$ & \cellcolor{golden!10} $\textbf{66.86\%}$ & \cellcolor{golden!10} $\textbf{34.88\%}$ & \cellcolor{golden!10} $\textbf{74.24\%}$ & \cellcolor{golden!10} $\textbf{53.02\%}$\\

\cline{2-10}\noalign{\smallskip}

& \multirow{3}{*}{Gowal \emph{et al.} \cite{gowal2020uncovering}} & \multirow{3}{*}{$L_{\infty}, \epsilon=8/255$}  & \multirow{3}{*}{WRN70-16} & Base & $\textbf{91.09\%}$ & $65.88\%$ & $25.95\%$ & $66.43\%$ & $27.21\%$\\

& & & & DRQ \cite{schwinn2022improving} & $90.77\%$ & $71.00\%$ & $35.89\%$ & $72.87\%$ & $39.51\%$\\

 & & & & \cellcolor{golden!10} \AlgoName  & \cellcolor{golden!10} $88.18\%$ & \cellcolor{golden!10} $\textbf{72.02}\%$ & \cellcolor{golden!10} $\textbf{40.30\%}$ & \cellcolor{golden!10} $\textbf{75.90\%}$ & \cellcolor{golden!10} $\textbf{49.21\%}$\\

\hline\noalign{\smallskip}

\multirow{9}{*}{CIFAR100} & \multirow{5}{*}{Rebuffi \emph{et al.} \cite{rebuffi2021fixing}} & \multirow{5}{*}{$L_{\infty}, \epsilon=8/255$} & \multirow{5}{*}{WRN28-10}  & Base & $\textbf{62.40}\%$ & $32.06\%$ & $12.47\%$ & $38.32\%$ & $18.86\%$\\

& & & & TTE \cite{perez2021enhancing} &  $62.35\%$ &  $33.25\%$ & $13.84\%$ & $39.14\%$ & $20.22\%$\\

& & & & DRQ \cite{schwinn2022improving} & $61.32\%$ & $38.22\%$ & $19.41\%$ & $44.58\%$ & $26.78\%$\\

& & & & \cellcolor{golden!10} \AlgoName & \cellcolor{golden!10} $55.18\%$ & \cellcolor{golden!10} $37.61\%$ & \cellcolor{golden!10} $19.72\%$ & \cellcolor{golden!10} $45.86\%$ & \cellcolor{golden!10} $32.79\%$\\

& & & & \cellcolor{golden!10} \AlgoNameTop  & \cellcolor{golden!10} $57.95\%$ & \cellcolor{golden!10} $\textbf{38.73\%}$ & \cellcolor{golden!10} $\textbf{19.87\%}$ & \cellcolor{golden!10} $\textbf{47.56\%}$ & \cellcolor{golden!10} $\textbf{33.01\%}$\\

\cline{2-10}\noalign{\smallskip}

& & & & Base & $\textbf{69.15\%}$ & $36.90\%$ & $13.64\%$ & $40.86\%$ & $17.20\%$\\

& & & & DRQ \cite{schwinn2022improving} & $69.12\%$ & $43.96\%$ & $20.25\%$ & $48.95\%$ & $25.43\%$\\

& & & & \cellcolor{golden!10} \AlgoName & $59.83\%$ & \cellcolor{golden!10} $44.66\%$ & \cellcolor{golden!10} $\textbf{23.81\%}$ & \cellcolor{golden!10} $51.32\%$ & \cellcolor{golden!10} $\textbf{37.02\%}$\\

& \multirow{-4}{*}{Gowal \emph{et al.} \cite{gowal2020uncovering}}  & \multirow{-4}{*}{$L_{\infty}, \epsilon=8/255$} & \multirow{-4}{*}{WRN70-16}  & \cellcolor{golden!10} \AlgoNameTop  & \cellcolor{golden!10} $62.47\%$ & \cellcolor{golden!10} $\textbf{46.09\%}$ & \cellcolor{golden!10} $23.48\%$ & \cellcolor{golden!10} $\textbf{53.06\%}$ & \cellcolor{golden!10} $36.19\%$ \\

\hline\noalign{\smallskip} \hline\noalign{\smallskip}

\end{tabular}
}
% \vspace{-20pt}
\label{table:cifar10_and_cifar100}
\end{center}
\end{table*}

\begin{figure}[h!]
    
  \begin{center}
    \includegraphics[width=0.4\textwidth]{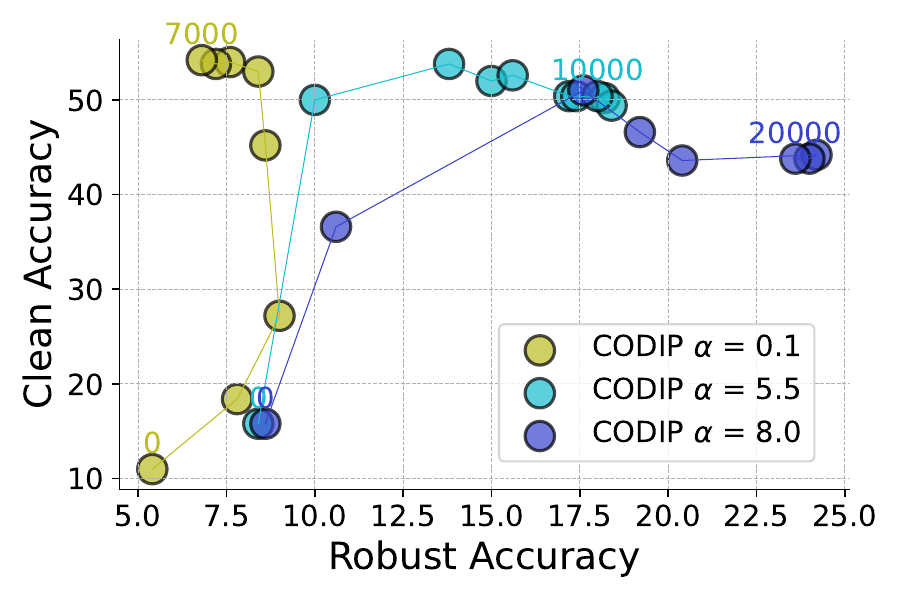}
  \end{center}
  \vspace{-15pt}
  \caption{
    \textbf{Impact of $\gamma$ on Clean-Robust Accuracy Trade-off } We present three $\alpha$ working points on the ImageNet dataset using an AT model $L_2,\epsilon = 3.0$.
    }
  \label{fig:gamma_tradeoff}
\end{figure}

\AlgoNameNoSpace's objective, which appears in \cref{alg:CODIP_calc_grad}, weights two targets: 
The first term, $L_{CE} \left(f \left(x + \mathbf{T}_{ij}\right), i \right)$, is a cross entropy loss between the classifier's prediction $f(\cdot)$ over the $j^{th}$ step of the $i^{th}$ transformation $x + \mathbf{T}_{ij}$, and the current class $i$. 
The goal of this term is to measure the classifier's error towards class $i$ and to transform the image so it semantically looks like class $i$.
The goal of the second term, $\gamma \| \mathbf{T}_{ij} \|_2 $, is to regularize the transformation to stay close to the input image $x$. 
This regularization is needed since we desire to change the input image toward class $i$ with minimal semantic changes.
Without this term, the transformation could turn the image into any instance of class $i$. 
Even in the case where $i$ is the true class, we can change the image into a completely different instance of the class, ultimately nullifying the efficacy of our method.
To conclude, balancing between the two terms, which is determined by the hyper-parameter $\gamma$, influences the class-similarity transformation balance, hence, it is essential.%for the success.% of the classification.
We demonstrate the impact of changing this value in \cref{fig:gamma_tradeoff}, and discuss it further in Appendix H.

\subsection{\AlgoNameTopNoSpace}
While \AlgoName is an effective algorithm that significantly enhances the classifier robustness, its inference time grows with the number of datasets' classes. 
To mitigate this issue, we propose \AlgoNameTopNoSpace, which reduces the inference time significantly by filtering out the classes with the lowest probability.

The classifier predicts the probability of the input image belonging to each class $f(x) \rightarrow \mathbb{R}^N \in [0,1]^N$. 
One way to speed up our algorithm is by limiting our prediction only to the most probable classes predicted by the classifier.
More specifically, we choose a group of $k$ classes $C_k \subset C=\{1, \dots, N\}$ containing the top-$k$ most probable prediction made by the classifier.
Next, we modify our method, which is presented in \cref{alg:CODIP}, by adapting \cref{alg:loop_over_classes} to transform only towards the most probable classes $C_k$.

\begin{table*}[hb!]
\begin{center}
\vspace{20pt}
\caption{
    \textbf{ImageNet and Flowers Results} 
        The performance of test-time methods over ImageNet and Flowers. The results are grouped per AT model, evaluating different test-time methods: `Base' in which we do not use any test-time method, followed by other methods.
}
\resizebox{\textwidth}{!}{%

\begin{tabular}{lllllccccc}
\hline\noalign{\smallskip}\hline

\rowcolor{gray!5}  & &  &  &  &  & \multicolumn{4}{c}{Attack}\\
\rowcolor{gray!5} & & & & & & \multicolumn{2}{c}{$L_{\infty}$} & \multicolumn{2}{c}{$L_{2}$} \\
\rowcolor{gray!5} \multirow{-3}{*}{Dataset} & \multirow{-3}{*}{AT Method} & \multirow{-3}{*}{\makecell{Trained Threat\\ Model}} & \multirow{-3}{*}{Architecture}  & \multirow{-3}{*}{\makecell{Test-Time \\ Method}}  & \multirow{-3}{*}{Clean} & $4/255$ & $8/255$ & $3.0$ & $6.0$\\

\hline\noalign{\smallskip}\hline\noalign{\smallskip}

\multirow{16}{*}{Imagenet} & \multirow{3}{*}{Madry et al. \cite{madry2017towards}} & \multirow{3}{*}{$L_{2}, \epsilon=3.0$}  & \multirow{3}{*}{RN50}  & Base & $57.90\%$ & $24.82\%$ & $05.26\%$ & $30.85\%$ & $10.35\%$\\

 & & & & TTE \cite{perez2021enhancing} & $\textbf{57.97}\%$ &  $26.56\%$ & $06.56\%$ & $32.33\%$ & $11.71\%$\\

 & & & & \cellcolor{golden!10} \AlgoNameTop &  \cellcolor{golden!10} $51.51\%$ & \cellcolor{golden!10} $\textbf{40.64\%}$ & \cellcolor{golden!10} $\textbf{22.10\%}$ & \cellcolor{golden!10} $\textbf{45.75\%}$ & \cellcolor{golden!10} $\textbf{34.0\%}$\\

\cline{2-10}\noalign{\smallskip}

 & \multirow{3}{*}{Salman et al. \cite{salman2020adversarially}} & \multirow{3}{*}{$L_{2}, \epsilon=3.0$} & \multirow{3}{*}{WRN50-2} & Base & $66.91\%$ & $30.86\%$ & $06.19\%$ & $38.27\%$ & $12.85\%$\\

 & & & & TTE \cite{perez2021enhancing} & $\textbf{66.94\%}$ &  - & $07.61\%$ & - & $14.57\%$\\

 & & & & \cellcolor{golden!10} \AlgoNameTop & \cellcolor{golden!10} $63.27\%$ & \cellcolor{golden!10} $\textbf{47.68\%}$ & \cellcolor{golden!10} $\textbf{25.23\%}$ & \cellcolor{golden!10} $\textbf{53.35\%}$ & \cellcolor{golden!10} $\textbf{38.36\%}$\\

\cline{2-10}\noalign{\smallskip}

&  \multirow{3}{*}{Madry et al. \cite{madry2017towards}}  & \multirow{3}{*}{$L_{\infty}, \epsilon=4/255$} & \multirow{3}{*}{RN50} & Base & $62.42\%$ & $28.96\%$ & $08.12\%$ & $07.06\%$ & $00.39\%$\\

&  & & & TTE \cite{perez2021enhancing} & $\textbf{62.58}\%$ &  $31.02\%$ & $09.69\%$ & $12.14\%$ & $01.25\%$\\

& & & & \cellcolor{golden!10}  \AlgoNameTop & \cellcolor{golden!10} $55.75\%$ & \cellcolor{golden!10} $\textbf{40.45\%}$ & \cellcolor{golden!10} $\textbf{17.97\%}$ & \cellcolor{golden!10} $\textbf{18.03\%}$ & \cellcolor{golden!10} $\textbf{04.42\%}$\\

\cline{2-10}\noalign{\smallskip}

& \multirow{3}{*}{Salman et al. \cite{salman2020adversarially}} & \multirow{3}{*}{$L_{\infty}, \epsilon=4/255$} & \multirow{3}{*}{WRN50-2} & Base & $68.41\%$ & $37.80\%$ & $12.84\%$ & $07.03\%$ & $00.21\%$\\

& & & & TTE \cite{perez2021enhancing} & $\textbf{68.61}\%$ & - & $13.32\%$ & - & $01.78\%$\\

& & & & \cellcolor{golden!10} \AlgoNameTop & \cellcolor{golden!10} $60.44\%$ & \cellcolor{golden!10} $\textbf{48.44\%}$ & \cellcolor{golden!10} $\textbf{25.44\%}$ & \cellcolor{golden!10} $\textbf{20.18\%}$ & \cellcolor{golden!10} $\textbf{04.24\%}$\\

% #################TRANS #######################
\cline{2-10}\noalign{\smallskip}

& \multirow{2}{*}{Debenedetti et al. \cite{debenedetti2023light}} & \multirow{2}{*}{$L_{\infty}, \epsilon=8/255$} & \multirow{2}{*}{XCiT-S} & Base & $\textbf{84.70}\%$ & $58.70\%$ & $32.60\%$  & $45.20\%$ & $14.50\%$ \\
  
&  &  & & \cellcolor{golden!10} \AlgoNameTop & \cellcolor{golden!10} $81.00\%$ & \cellcolor{golden!10} $\textbf{67.30}\%$ & \cellcolor{golden!10} $\textbf{44.00}\%$  & \cellcolor{golden!10} $\textbf{60.00}\%$ & \cellcolor{golden!10} $\textbf{27.50}\%$ \\

\cline{2-10}\noalign{\smallskip}

& \multirow{2}{*}{Debenedetti et al. \cite{debenedetti2023light}} & \multirow{2}{*}{$L_{\infty}, \epsilon=8/255$} &  \multirow{2}{*}{XCiT-M} &  Base & $\textbf{89.55}\%$ & $63.80\%$ & $33.70\%$  & $45.70\%$ & $11.20\%$ \\
& & &  & \cellcolor{golden!10} \AlgoNameTop & \cellcolor{golden!10} $87.70\%$ & \cellcolor{golden!10} $\textbf{72.80}\%$ & \cellcolor{golden!10} $\textbf{45.70}\%$  & \cellcolor{golden!10} $\textbf{61.40}\%$ & \cellcolor{golden!10} $\textbf{22.10}\%$ \\

\hline\noalign{\smallskip}

\multirow{2}{*}{Flowers} & \multirow{2}{*}{Debenedetti et al. \cite{debenedetti2023light}} & \multirow{2}{*}{$L_{\infty}, \epsilon=8/255$} & \multirow{2}{*}{XCiT-S} &  Base & $\textbf{79.30}\%$ & $62.40\%$ & $39.60\%$  & $55.60\%$ & $23.20\%$ \\
& &  & & \cellcolor{golden!10} \AlgoNameTop & \cellcolor{golden!10} $78.30\%$ & \cellcolor{golden!10} $\textbf{69.80}\%$ & \cellcolor{golden!10} $\textbf{50.40}\%$  & \cellcolor{golden!10} $\textbf{68.40}\%$ & \cellcolor{golden!10} $\textbf{45.70}\%$ \\

\hline\noalign{\smallskip}\hline\noalign{\smallskip}

\end{tabular}
}
% \vspace{-20pt}
\label{table:imagenet}
\end{center}
\end{table*}

\section{Experiments}
\label{sec:experiments}
In this section, we provide implementation details, along with qualitative and quantitative evidence supporting our method. 
We present the datasets, baselines, and implementation details, followed by the results on CIFAR-10, CIFAR-100 \cite{krizhevsky2009learning}, ImageNet \cite{deng2009imagenet}, and Flowers \cite{nilsback2008automated}. 
We compare our method using common practice attacks, including black-box attacks. 
Next, we introduce a method to control the clean-robust accuracy trade-off and conclude with a thorough runtime comparison.

In the Appendix, we include an ablation study of each component of our method in Appendix A, qualitative results in Appendix B, white-box attack in Appendix C, the influence of a designated Top-$k$ attack in Appendix D, further implementation details in Appendix E and further explanations in Appendix F, Appendix G, Appendix H.

\subsection{Experimental Settings}
\paragraph{\normalfont\textbf{Datasets}} The dataset CIFAR10 and CIFAR100 \cite{krizhevsky2009learning} both containing $50\text{K}$ samples for training and $10\text{K}$ samples for test, each of size $(3 \times 32 \times 32)$. 
They differ in the number of classes, where CIFAR10 contains $10$ classes and CIFAR100 contains $100$. 
The ImageNet dataset \cite{deng2009imagenet} contains $1\text{M}$ images for training and $50\text{K}$ for validation, the images are resized to $(3 \times 256 \times 256)$ and cropped to $(3 \times 224 \times 224)$.
The Flowers dataset \cite{nilsback2008automated} contains $103$ classes, $1030$ images for training and $6129$ for test, the images are resized to $(3 \times 256 \times 256)$ and cropped to $(3 \times 224 \times 224)$.

\paragraph{\normalfont\textbf{Evaluation}} For evaluation we use the test set of CIFAR10, CIFAR100 and Flowers, and the validation set of ImageNet as we follow the convention of ImageNet evaluation, and we report the average accuracy. 
We employ AutoAttack \cite{croce2020reliable} as a robust and widely accepted benchmark for adversarial attack. 
AutoAttack is an ensemble of four diverse attacks, where all four attacks are applied to perform a successful attack and allow a reliable robustness evaluation. 
It includes the attacks: APGD-CE and APGD-DLR which were proposed in \cite{croce2020reliable}, FAB \cite{croce2020minimally} and Square Attack \cite{andriushchenko2020square}. 
For CIFAR10 and CIFAR100 we evaluate the performance using the following thereat models $(\ell_{\infty}, \epsilon=8/255)$, $(\ell_{\infty}, \epsilon=16/255)$, $(\ell_{2}, \epsilon=0.5)$, $(\ell_{2}, \epsilon=1.0)$, and for ImageNet and Flowers we use $(\ell_{\infty}, \epsilon=4/255)$, $(\ell_{\infty}, \epsilon=8/255)$, $(\ell_{2}, \epsilon=3.0)$, $(\ell_{2}, \epsilon=6.0)$.
Following previous works \cite{laidlaw2020perceptual, hill2020stochastic, yoon2021adversarial}, we attack on the classifier, as it is computationally impossible to attack these methods. 

We follow the common data split and portion, proposed by Trustworthy-AI-Group\footnote{https://github.com/Trustworthy-AI-Group/TransferAttack}, for all of the transformer based models and for the black-box attacks.

\paragraph{\normalfont\textbf{Implementation Details}}
We check our method performance on SOTA AT trained classifiers.
For CIFAR10 and CIFAR100 we use Rebuffi \emph{et al.} \cite{rebuffi2021fixing}, Gowal \emph{et al.} \cite{gowal2020uncovering} and Madry \emph{et al.} \cite{madry2017towards}, for ImageNet we use Madry \emph{et al.}, Salman \emph{et al.} \cite{salman2020adversarially} and Debenedetti et al. \cite{debenedetti2023light}, and for Flowers we use Debenedetti et al. \cite{debenedetti2023light}.
For each AT classifier, we specify the chosen hyper-parameters in Appendix E, and for each transformation we perform $30$ iterations. 
We use the following architectures: ResNet50, WideResNet50-2, WideResNet28-10, WideResNet70-16, XCiT-S/M.

\paragraph{\normalfont\textbf{Baselines}} 
To evaluate our method we compare it to a few baselines.
We start with `Base' which is the AT classifier and PAT \cite{laidlaw2020perceptual} which suggests an AT threat model-agnostic defense. 
PAT \cite{laidlaw2020perceptual} and Randomized Smoothing \cite{cohen2019certified} were evaluated only on the CIFAR10 dataset.
Additionally, we compare our method to two other test-time methods: TTE \cite{perez2021enhancing} and DRQ \cite{schwinn2022improving}. 
We evaluate both of these methods on different threat models, allowing a fair comparison.
TTE was evaluated on CIFAR10, CIFAR100, and ImageNet, however, DRQ can not be applied to ImageNet, as stated in their work, since it is computationally infeasible.

We propose two variants of our method: method \AlgoName and \AlgoNameTopNoSpace.
Both are presented in \cref{sec:our_method} and while \AlgoName performs conditioned transformation of the input image towards all of the dataset's classes, \AlgoNameTop transforms only toward the group of classes that is more likely to contain the true class. 
\AlgoNameTop is essential when employing our method on datasets with a large number of classes.
To ensure a fair comparison, we explore a designated Top-$k$ attack which was proposed by Zhang \emph{et al.} \cite{zhang2022investigating} and we discuss our findings in Appendix D.

\subsection{Comparison}
% \vspace{20pt}

We present the performance over four datasets: 
CIFAR10 and CIFAR100 in \cref{table:cifar10_and_cifar100}, and ImageNet and Flowers in \cref{table:imagenet}.
We present AT models that were trained on different Trained Threat Model and with different architecture, and we employ different test-time method: Base, TTE DRQ, \AlgoName and \AlgoNameTop. 
For CIFAR10 and CIFAR100 datasets we compare our method to PAT \cite{laidlaw2020perceptual}, Randomized Smoothing \cite{cohen2019certified}, TTE \cite{perez2021enhancing} and to DRQ \cite{schwinn2022improving}, and for ImageNet we use TTE \cite{perez2021enhancing}.

\AlgoName enhances the base AT methods (represented by `Base' under `Test-Time Method') for both seen and unseen attacks. 
In particular, we enhance the performance for seen attacks by up to $9\%$, $10\%$, $15\%$, $11\%$, and 
for unseen attacks by up to $10\%$, $20\%$, $26\%$, $22\%$ over CIFAR10, CIFAR100, Imagenet, and Flowers respectively.

When comparing to other test-time methods, we enhance the performance for seen attacks by up to $1\%$, $3\%$ and $12\%$, and for unseen attacks by up to $10\%$, $20\%$ and $24\%$ over CIFAR10, CIFAR100 and Imagenet respectively.
Additionally, for CIFAR10 we compare \AlgoName to the PAT \cite{laidlaw2020perceptual}, a method that addresses the issue of robustness to unseen attacks, and we report up to $52\%$ higher accuracy.
We further compare our method to Randomized Smoothing \cite{cohen2019certified}, a test-time method requiring the robust classifier to be trained on Gaussian noise. 
Typically, robust classifiers are not trained using this augmentation, which often leads to ineffective outcomes with Randomized Smoothing, particularly with the models we employ, such as AT \cite{madry2017towards} RN50.
Moreover, for CIFAR100 we use both \AlgoName and \AlgoNameTop ($k=10$) demonstrating that not only do we benefit from much faster inference time but also that the performances are slightly better.
For ImageNet we use \AlgoNameTop ($k=20$), as \AlgoName is computationally not feasible, and for Flowers we use \AlgoNameTop ($k=10$).

\paragraph{\normalfont\textbf{Top-$k$}} The Top-$k$ version of CODIP sometimes outperforms the full CODIP because it focuses on the most likely classes, reducing the influence of less relevant or noisy class transformations. By concentrating on a smaller set of high-confidence predictions, the Top-$k$ version can enhance accuracy by avoiding the potential noise introduced when evaluating all possible classes. This selective attention allows it to maintain or even improve performance in scenarios where the correct class is consistently within the top $k$ predictions. Additional information in Appendix F.

\setlength{\tabcolsep}{4pt}
\begin{table}[h!]
\begin{center}
\caption{
    \textbf{Black-Box Attacks} Two different Trained Threat Models (TTM) are evaluated. 
    Both RN50-2 models were adversarially trained by Salman et al. \cite{salman2020adversarially}.
    Each model is evaluated using Imagenet \cite{deng2009imagenet} dataset, under clean images, three black-box attacks, and compared to the base model.
    }
    
\resizebox{\columnwidth}{!}{%

\begin{tabular}{llcccc}
\hline\noalign{\smallskip}\hline
\rowcolor{gray!5} & & & & & \\
\rowcolor{gray!5} \multirow{-2}{*}{TTM}  & \multirow{-2}{*}{\makecell{Test Time\\ Method}} & \multirow{-2}{*}{Clean} & \multirow{-2}{*}{MI-FGSM \cite{dong2018boosting}} & \multirow{-2}{*}{Admix \cite{wang2021admix}} & \multirow{-2}{*}{L2T \cite{zhu2024learning}}\\

\hline\noalign{\smallskip}\hline\noalign{\smallskip}
 \multirow{2}{*}{$L_2$, $\epsilon=3.0$ } & Base & $\textbf{89.40}\%$ & $79.70\%$ & $74.20\%$  & $73.10\%$ \\
 
 & \AlgoNameTop & $87.80\%$ & $\textbf{80.60}\%$ & $\textbf{76.40}\%$  & $\textbf{74.30}\%$ \\

 \multirow{2}{*}{ $L_{\infty}$, $\epsilon=4/255$ } & Base & $\textbf{91.10}\%$ & $83.40\%$ & $78.30\%$  & $76.30\%$\\
 & \AlgoNameTop & $91.00\%$ & $\textbf{84.70}\%$ & $\textbf{80.00}\%$  & $\textbf{78.30}\%$ \\

\hline\noalign{\smallskip}\hline\noalign{\smallskip}

\end{tabular}

}
\label{table:black_box}

\end{center}
\end{table}

\subsection{Black Box Attack}
\label{exp:black_box}

We present the effectiveness of our method against advanced black box attacks in \cref{table:black_box}. 
including MI-FGSM \cite{dong2018boosting} and Admix \cite{wang2021admix}, and L2T \cite{zhu2024learning} which is currently considered the strongest one. 
We utilize the official implementation Trustworthy-AI-Group\footnote{https://github.com/Trustworthy-AI-Group/TransferAttack}, maintain the same data split, and use an ensemble of surrogate models under the $L_{\infty}=16/255$ attack. As demonstrated, our defense enhances the performance under black-box attacks as well.

\begin{table*}[!ht]
\begin{center}
\caption{
    \textbf{Runtime and Compute} Comparing Base, TTE \cite{perez2021enhancing}, DRQ \cite{schwinn2022improving}, \AlgoName, and \AlgoNameTopNoSpace. 
    For both \AlgoName and $\text{CODIP}_{\text{Top-}5}$ we provide the runtime alongside Maximum Batch-Size (MBS).
    }
\resizebox{\textwidth}{!}{%

\begin{tabular}{lllccccccc}
\hline\noalign{\smallskip}\hline

\rowcolor{gray!5} & & & & & & & & \\
\rowcolor{gray!5} \multirow{-2}{*}{Dataset} & \multirow{-2}{*}{Method} & \multirow{-2}{*}{Architecture} & \multirow{-2}{*}{Base} & \multirow{-2}{*}{TTE \cite{perez2021enhancing}} & \multirow{-2}{*}{DRQ \cite{schwinn2022improving}} & \multirow{-2}{*}{\makecell{\AlgoName \\ Times / MBS}} & \multirow{-2}{*}{$\text{CODIP}_{\text{Top-}20}$} & \multirow{-2}{*}{$\text{CODIP}_{\text{Top-}10}$} & \multirow{-2}{*}{\makecell{$\text{CODIP}_{\text{Top-}5}$ \\ Times / MBS}}  \\

\hline\noalign{\smallskip}\hline\noalign{\smallskip}

\multirow{3}{*}{CIFAR10} & AT \cite{madry2017towards} & RN50 & $\times 1$ & $\times 10$ &  $\times 1,192$ & $\times 144 / 150$   & $-$ & $-$ & $\times 144 / 300$  \\ 
 & Rebuffi et al. \cite{rebuffi2021fixing} & WRN28-10 & $\times 1$ & $\times 10$ & $\times 1,441$ & $\times 126 / 128$ & $-$ & $-$ & $\times 119 / 256 $ \\
 & Gowal et al. \cite{gowal2020uncovering} & WRN70-16 & $\times 1$ & $\times 11$ & $\times 1,510$ & $\times 240 / 36$  & $-$ & $-$ & $\times 143 / 72$ \\

\hline\noalign{\smallskip}

\multirow{2}{*}{CIFAR100} & Rebuffi et al.\cite{rebuffi2021fixing} &  WRN28-10 & $\times 1$ & $\times 10$ & $\times 14,128$ & $\times 1,017 / 12$ & $\times  205$ & $\times 123$  & $\times 112 / 280$ \\  

 & Gowal et al. \cite{gowal2020uncovering} &  WRN28-10 & $\times 1$ & $\times 11$ & $\times 15,193 $ & $\times 1,776 / 3$  & $\times  420$ & $\times 246$ & $\times 146 / 65$ \\

\hline\noalign{\smallskip}

\multirow{1}{*}{ImageNet} & AT \cite{madry2017towards}&  RN50 & $\times 1$ & $\times 7$ & $-$ & $-$ & $\times 133$  & $\times 90$ & $\times 90 / 80$ \\ % & $\times 20 / 40$  \\

\hline\noalign{\smallskip}\hline\noalign{\smallskip}

\end{tabular}

}
\label{table:runtime}
\vspace{-15pt}

\end{center}
\end{table*}

\subsection{Clean-Robust Accuracy Trade-off}
% \vspace{20pt}
The clean-robust accuracy trade-off is a well-known phenomenon \cite{zhang2019theoretically, tsipras2018robustness}. 
Training a classifier with adversarially perturbed images can significantly improve the robust accuracy, but it often comes at the expense of clean accuracy. 
When robustifying a model using AT, this trade-off is determined during training and cannot be controlled afterwards. 
Similarly, test-time methods also operate with a fixed clean-robust accuracy trade-off and do not offer a way to control it. Further details are provided in Appendix G.

\begin{figure}[h!]
    
  \begin{center}
    \includegraphics[width=0.5\textwidth]{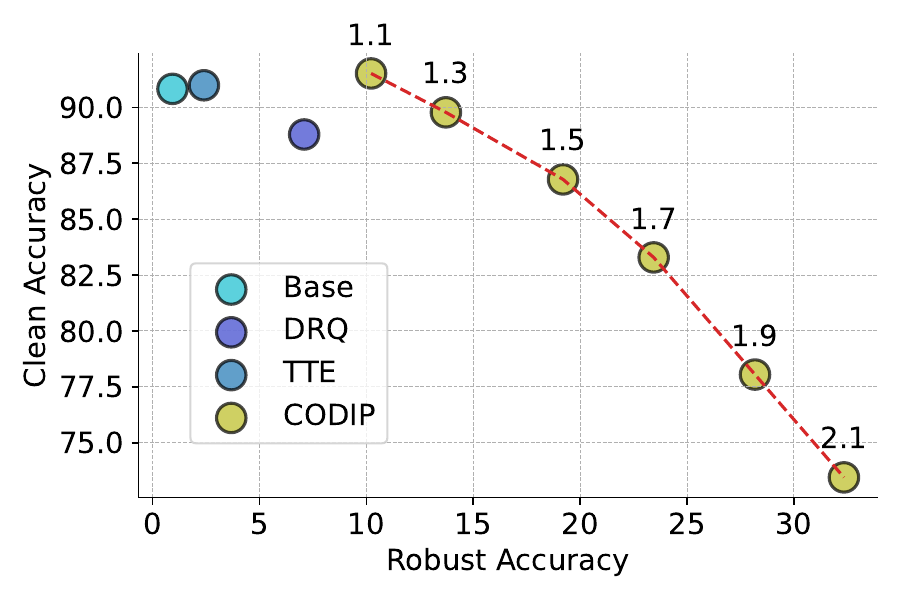}
  \end{center}
  \caption{
    \textbf{Clean-Robust Accuracy Trade-off} A demonstration of our proposed controlled clean-robust accuracy tradeoff. 
    The tradeoff is controlled by adapting the step size value $\alpha$, specified beside each of \AlgoName workpoints.
    The used test-time methods: `Base' which is the base AT model, TTE \cite{perez2021enhancing}, DRQ \cite{schwinn2022improving}, and \AlgoNameNoSpace.
}
  \label{fig:tradeoff}
\end{figure}

Our method, on the other hand, allows the user to control the trade-off by adjusting $\alpha$, which is the transformation's step size, presented in \cref{alg:CODIP}. 
By increasing $\alpha$, the user can prioritize robust accuracy, while decreasing $\alpha$ prioritize clean accuracy. 
In \cref{fig:tradeoff}, we present the clean-robust accuracy tradeoff for AT model trained on CIFAR10 by Madry et al.'s \cite{madry2017towards} on threat model $L_{2}, \epsilon=0.5$ and architecture ResNet-50.
We compare the performance of the robust classifier under the attack threat model ($L_{\infty}, \epsilon=16/255$) using the AT model and three other test-time methods: Base which is the performance of the AT classifier, TTE \cite{perez2021enhancing}, DRQ \cite{schwinn2022improving}, and our method, \AlgoName. 
As depicted in the figure, \AlgoName we can control the trade-off by adjusting $\alpha$, prioritizing clean or robust accuracy as needed.
We show that we can improve the clean accuracy or enhance robust accuracy by up to $25\%$.

\subsection{Runtime Analysis}
\label{exp:runtime}

This section provides a runtime comparison across different architectures and datasets, detailed in \cref{table:runtime}.
For a fair comparison, all experiments were conducted using one RTX A6000 with a batch size of $1$.

We compare our method to the base classifier, TTE \cite{perez2021enhancing} and to DRQ \cite{schwinn2022improving}, reporting the relative inference time compared to the base classifier.
We can see that TTE \cite{perez2021enhancing} increase the inference time due to the predetermined number of augmentations performed for each image.
This extra time is added since we perform a predetermined number of augmentation for each image. 
The other methods, DRQ \cite{schwinn2022improving} and our approach, also result in increased inference time compared to the base classifier. 
However, both \AlgoName and, in particular, \AlgoNameTop demonstrate significantly faster performance. Specifically, \AlgoNameTop is up to 100 times faster than DRQ \cite{schwinn2022improving}.

Another crucial consideration is memory consumption, detailed under both \AlgoName and $\text{CODIP}_{\text{Top-}5}$ columns.
We specify the Maximum Batch Size (MBS) that fits into memory.
It is evident that $\text{CODIP}_{\text{Top-}5}$ enables an increase in batch size of up to $23$ times.

% \section{Ablation Study}
% \label{sec:ablation}

% We perform an ablation study in \cref{table:ablation} at \cref{appendix:ablation}, evaluating the influence of different elements on the performance.
% We use Rebuffi \emph{et al.} \cite{rebuffi2021fixing} AT model WRN28-10 trained on CIFAR10 and threat model $\ell_{\infty}=8/255$.
% We evaluate the performance over clean examples and 4 attacks: $(\ell_{\infty}, \epsilon=8/255)$, $(\ell_{\infty}, \epsilon=16/255)$, $(\ell_{2}, \epsilon=0.5)$, $(\ell_{2}, \epsilon=1.0)$. 
% First, we investigate the necessity of the AT model.
% Next, we check the impact of the distance metric by which we make the prediction.
% Moreover, we check the influence of the transformation hyper-parameters $\alpha, \gamma$, and transformation steps $M$.
% Finally, we check different $k$ values for \AlgoNameTopNoSpace.

\section{Conclusion}
\label{sec:conclusion}

In this work, we introduce \AlgoName -- a novel test-time threat model-agnostic method that enhances AT models through conditional image transformation and distance-based prediction.
Our method leverages AT models specialized in a specific threat model, boosts their performance on the trained threat model, and equips them with threat model-agnostic capabilities without further training or access to the training data.   
Additionally, we provide a controllable, clean-robust accuracy trade-off mechanism that can be adapted by the user as required.
Moreover, we propose \AlgoNameTop, an efficient method that speeds up the inference time.
Finally, we conducted an extensive evaluation to demonstrate state-of-the-art results on various architectures, convolution or transformer based, AT training methods, white-box and black-box attacks, and four widely used datasets: CIFAR10, CIFAR100, ImageNet, and Flowers.

\section*{Acknowledgments}
This project has received funding from the European Research Council (ERC) under the European Union’s Horizon 2020 research and innovation programme (grant agreement No. 863839).
This research was partially supported by NOFAR MAGNET number 78732 by the
Israel Innovation Authority.

%%%%%%%%% REFERENCES
{\small
\bibliographystyle{ieee_fullname}
\bibliography{egbib}
}

% \clearpage
% \input{appendix}

\end{document}

\title{Supplementary Material for Class-Conditioned Transformation for Enhanced Robust Image Classification}

\author{
    Tsachi Blau\textsuperscript{\dag}\thanks{\texttt{tsachiblau@campus.technion.ac.il}} \quad 
    Roy Ganz\textsuperscript{\dag} \quad 
    Chaim Baskin\textsuperscript{\ddag} \quad 
    Michael Elad\textsuperscript{\dag} \quad 
    Alex M. Bronstein\textsuperscript{\dag}
}

\maketitle
\setcounter{page}{11}
\renewcommand{\thefootnote}{\fnsymbol{footnote}}
\footnotetext[2]{Department of Computer Science Technion - Israel Institute of Technology}
\footnotetext[3]{School of Electrical and Computer Engineering - Ben-Gurion University of the Negev}

\section{Ablation Study}
\label{app:ablation}

We perform an ablation study, evaluating the influence of different elements on the performance.
We use Rebuffi \emph{et al.} \cite{rebuffi2021fixing} AT model WRN28-10 trained on CIFAR10 and threat model $\ell_{\infty}=8/255$.
We evaluate the performance over clean examples and 4 attacks: $(\ell_{\infty}, \epsilon=8/255)$, $(\ell_{\infty}, \epsilon=16/255)$, $(\ell_{2}, \epsilon=0.5)$, $(\ell_{2}, \epsilon=1.0)$. 
First, we investigate the necessity of the AT model.
Next, we check the impact of the distance metric by which we make the prediction.
Moreover, we check the influence of the transformation hyper-parameters $\alpha, \gamma$, and transformation steps $M$.
Finally, we check different $k$ values for \AlgoNameTopNoSpace.

\paragraph{\normalfont\textbf{PAG Property}}
Our method is based on image transformations, and we show the effectiveness of the proposed transformation through extensive evaluation.
We claim that the transformation is possible due to the PAG property, which is known to be possessed by AT classifiers \cite{engstrom2019adversarial, etmann2019connection,ross2018improving,tsipras2018robustness}. 
To validate this claim we use our method on WRN28-10 classifier that was trained on clean images.
We show that the model enhancement is limited compared to the results achieved by AT models.

\setlength{\tabcolsep}{4pt}
\begin{table*}[ht!]
\begin{center}
\caption{
\textbf{Ablation Study} An ablation over (a) The necessity of the AT model, (b) different distance metrics, (c) different hyper-parameters values, and (d) different $k$ values.
    % \vspace{-30pt}
}
\resizebox{0.9\textwidth}{!}{%

\begin{tabular}{cccccccccccc}

\hline\noalign{\smallskip}\hline

\multirow{3}{*}{AT} & \multirow{3}{*}{Distance} & \multirow{3}{*}{\makecell{Transformation Steps \\ $M$}} & \multirow{3}{*}{$\alpha$} & \multirow{3}{*}{$\gamma$} & \multirow{3}{*}{Top-k} &  \multirow{3}{*}{Clean} & \multicolumn{4}{c}{Attack}\\
& & & & & & & \multicolumn{2}{c}{$L_{\infty}$} & \multicolumn{2}{c}{$L_{2}$}\\
& & & & & & &  $8/255$ & $16/255$ & $0.5$ & $1.0$\\

\hline\noalign{\smallskip}\hline\noalign{\smallskip}

\multirow{2}{*}{\xmark} & $-$ & $-$ & $-$ & $-$ & $-$ & $\textbf{95.26\%}$ & $00.00\%$ & $00.00\%$ & $00.00\%$ & $00.00\%$ \\
 & $\ell_2$ & $30$ & $0.05$ & $300$ & \xmark & $93.04\%$ & $\textbf{01.11\%}$ & $\textbf{01.12\%}$ & $\textbf{01.24\%}$ & $\textbf{01.10\%}$ \\

\hline\noalign{\smallskip}

% \multirow{2}{*}{\cmark} & $-$ & $-$ & $-$ & $-$ & $-$ & $87.33\%$ & $60.77\%$ & $25.44\%$ & $66.72\%$ & $35.01\%$ \\
%  & $\ell_2$ & 30 & $0.1$ & $300$ & \xmark & $84.86\%$ & $66.96\%$ & $35.15\%$ & $74.84\%$ & $53.32\%$ \\

% \hline\noalign{\smallskip}

\multirow{3}{*}{\cmark} & LPIPS \cite{zhang2018unreasonable} & \multirow{2}{*}{30} & $1.5$ & $400$ & \multirow{2}{*}{\xmark} & $80.97\%$ & $66.49\%$ & $33.54\%$ & $74.73\%$ & $\textbf{59.25\%}$ \\
 & $\ell_2$ &  & $0.1$ & $300$ &  & $\textbf{84.86\%}$ & $\textbf{66.96\%}$ & $\textbf{35.15\%}$ & $\textbf{74.84\%}$ & $53.32\%$ \\

\hline\noalign{\smallskip}

\multirow{5}{*}{\cmark} & \multirow{5}{*}{$\ell_2$} & $100$ & $0.05$ & \multirow{5}{*}{$300$} & \multirow{5}{*}{\xmark} & $\textbf{84.86\%}$ & $66.91\%$ & $35.02\%$ & $\textbf{74.84\%}$ & $\textbf{53.50\%}$ \\
 &  & \multirow{4}{*}{30} & $0.05$ &  &  & $82.96\%$ & $65.40\%$ & $33.29\%$ & $74.14\%$ & $52.74\%$ \\
 &  &  & $0.1$ &  &  & $\textbf{84.86\%}$ & $66.96\%$ & $35.15\%$ & $\textbf{74.84\%}$ & $53.32\%$ \\
 &  &  & $0.5$ &  &  & $82.14\%$ & $\textbf{67.85\%}$ & $\textbf{39.11\%}$ & $73.32\%$ & $52.25\%$ \\
 &  &  & $1.0$ &  &  & $73.78\%$ & $63.78\%$ & $32.39\%$ & $68.52\%$ & $47.10\%$ \\

\hline\noalign{\smallskip}

\multirow{5}{*}{\cmark} & \multirow{5}{*}{$\ell_2$} & \multirow{5}{*}{30} & \multirow{5}{*}{$0.1$} & 10 & \multirow{5}{*}{\xmark} & $34.01\%$ & $33.84\%$ & $29.76\%$ & $35.29\%$ & $41.65\%$ \\
 &  &  &  & 100 &  & $81.33\%$ & $\textbf{67.13\%}$ & $\textbf{38.47\%}$ & $74.02\%$ & $\textbf{57.05\%}$ \\
 &  &  &  & 300 &  & $\textbf{84.86\%}$ & $66.96\%$ & $35.15\%$ & $\textbf{74.84\%}$ & $53.32\%$ \\
 &  &  &  & 500 &  & $84.58\%$ & $66.08\%$ & $32.78\%$ & $73.94\%$ & $49.42\%$ \\
 &  &  &  & 1000 &  & $79.47\%$ & $63.63\%$ & $29.80\%$ & $69.46\%$ & $42.41\%$ \\

\hline\noalign{\smallskip}

\multirow{4}{*}{\cmark} & \multirow{4}{*}{$\ell_2$} & 10 & \multirow{4}{*}{$0.1$} & \multirow{4}{*}{$300$} & \multirow{4}{*}{\xmark} & $70.28\%$ & $61.67\%$ & $29.66\%$ & $69.32\%$ & $49.21\%$ \\
&  & 30 &  &  &  & $84.86\%$ & $66.96\%$ & $\textbf{35.15\%}$ & $74.84\%$ & $53.32\%$ \\
&  & 50 &  &  &  & $84.87\%$ & $66.98\%$ & $35.02\%$ & $74.76\%$ & $53.19\%$ \\
&  & 100 &  &  &  & $\textbf{84.88\%}$ & $\textbf{67.01\%}$ & $35.13\%$ & $\textbf{74.85\%}$ & $\textbf{53.34\%}$ \\

\hline\noalign{\smallskip}

\multirow{4}{*}{\cmark} & \multirow{4}{*}{$\ell_2$} & \multirow{4}{*}{30} & \multirow{4}{*}{$0.1$} & \multirow{4}{*}{$300$} & 2 & $\textbf{86.37\%}$ & $66.45\%$ & $32.70\%$ & $73.98\%$ & $49.67\%$ \\
 &  &  &  &  & 5 & $85.79\%$ & $\textbf{67.12\%}$ & $34.72\%$ & $\textbf{75.14\%}$ & $53.30\%$ \\
 &  &  &  &  & 7 & $85.63\%$ & $\textbf{67.12\%}$ & $\textbf{34.97\%}$ & $75.12\%$ & $\textbf{53.55\%}$ \\
 &  &  &  &  & 9 & $85.21\%$ & $66.97\%$ & $34.72\%$ & $74.91\%$ & $53.31\%$ \\

\hline\noalign{\smallskip}\hline\noalign{\smallskip}
\end{tabular}%
}
% \caption{
%     \textbf{Ablation Study} An ablation over 
% (a) The necessity of the AT model, (b) different distance metrics, (c) different hyper-parameters values, and (d) different $k$ values.
%     % \vspace{-30pt}
% }
\label{table:ablation}
\end{center}
\end{table*}

\paragraph{\normalfont\textbf{Distance Metric}}
Our method operates through two phases: transformation and distance measurement, where the distance is used for the classification decision. 
\AlgoName operates through $\ell_2$ norm, however, there are other reasonable choices such as LPIPS \cite{zhang2018unreasonable}, which measures the perceptual similarity between two images.
We demonstrate that $\ell_2$ better suit our method. 

\paragraph{\normalfont\textbf{Hyper-Parameters}}
We investigate the influence of each hyper-parameter, $\alpha, \gamma, M$, of \AlgoName over the model performance.
First, we look at different values of $\alpha$ which determines the transformation step size. 
The proposed transformation is performed via iterative gradient updates, where each step is of size $\alpha$.
Hence, the step size holds an important key. 
While small steps might better follow the function, the progress is slower since it requires additional steps.
On the other hand a large step size, although much faster,  might lead to insufficient results. 
Our results in \cref{table:ablation} support the theory, as a large step size leads to bad results.
While small step size, with additional transformation steps, leads to equally good performance.

Next, we examine the influence of $\gamma$ on the performance.
This parameter regulates the transformation distance.
Small values allow the transformation to change the image, while large values restrict the transformation to be minimal. 

Finally, we examine the number of transformation steps. 
When the number of steps is small, the transformation can not reach its objective.
For a sufficient number of steps, we get good results, even when we have more than the minimal required number of steps.  

\paragraph{\normalfont\textbf{Top-$k$}}
We investigate the influence of different $k$ values on \AlgoNameTop.
For small values, our method performance is getting closer to the AT classifier performances as we rely on its predictions.
When we increase $k$, we rely more on the transformation performance, which leads to better robustness.

\clearpage

\section{Qualitative Results}
\label{app:qualitative}

\begin{figure*}[h!]
    \centering
    \includegraphics[width=1.0\textwidth]{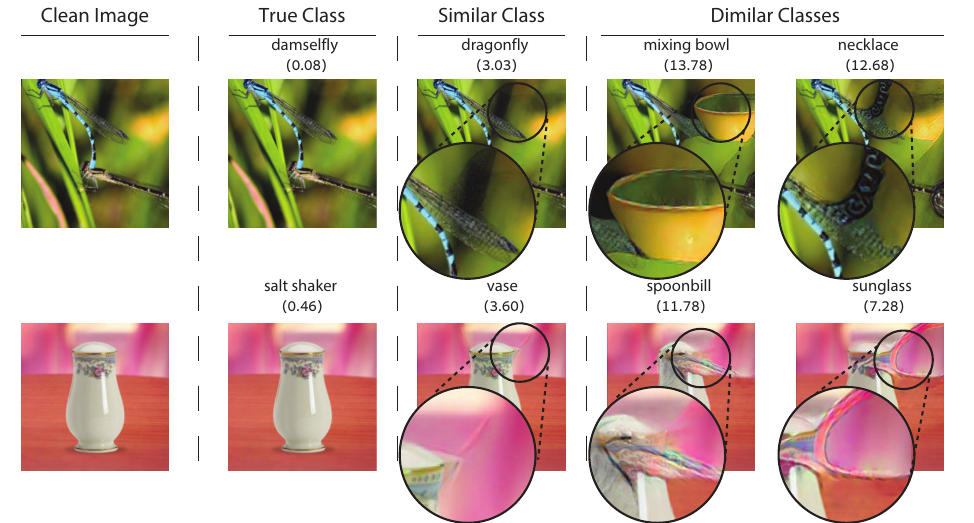}
    \caption{
    \textbf{Class-Conditioned Transformation} Depiction of class-conditioned transformations of clean images from ImageNet towards true class, similar class, and dissimilar class. 
    These transformations performed using \AlgoNameNoSpace, and the $\ell_2$ distance between the clean image and the transformed images is noted below the target class name.
    }
    \label{fig:generation_of_classes}
\end{figure*}

We perform a qualitative experiment in which we compare the transformation's visualization towards different classes.
In \cref{fig:generation_of_classes} we demonstrate that the transformation towards the true class does not change the image appearance much, while transforming to another class does.
The transformation to other classes changes the image considerably, especially towards classes that are not semantically similar to the true class, which supports the assumption underlying our method.

\clearpage

\section{White-Box Attack}
\label{app:white_box}

\begin{table}[hb!]
\begin{center}
\caption{\textbf{White-Box Attack} An white-box attack evaluation of over CIFAR10 dataset. 
We compare `Base', which is the AT classifier, to \AlgoNameTopNoSpace, using PGD attack \cite{madry2017towards}.
}
\resizebox{\textwidth}{!}{%

\begin{tabular}{llllccccc}
\hline\noalign{\smallskip}\hline

\rowcolor{gray!5}  &  & &  &  & \multicolumn{4}{c}{Attack}\\
\rowcolor{gray!5} & & & & & \multicolumn{2}{c}{$L_{\infty}$} & \multicolumn{2}{c}{$L_{2}$} \\
\rowcolor{gray!5} \multirow{-3}{*}{AT Method} & \multirow{-3}{*}{\makecell{Trained Threat\\ Model}} & \multirow{-3}{*}{Architecture}  & \multirow{-3}{*}{\makecell{Test-Time \\ Method}}  & \multirow{-3}{*}{Clean} & $8/255$ & $16/255$ & $0.5$ & $1.0$\\

\hline\noalign{\smallskip}\hline\noalign{\smallskip}

\multirow{2}{*}{Rebuffi \emph{et al.} \cite{rebuffi2021fixing}} & \multirow{2}{*}{$L_{\infty}, \epsilon=8/255$} & \multirow{2}{*}{WRN28-10}  & Base & $\textbf{87.33}\%$ & $64.40\%$ & $33.60\%$ & $70.39\%$ & $45.78\%$\\

 & & & \cellcolor{golden!10} \AlgoNameTopNoSpace & \cellcolor{golden!10} $84.38\%$ & \cellcolor{golden!10} $\textbf{73.13}\%$ & \cellcolor{golden!10} $\textbf{62.38}\%$ & \cellcolor{golden!10} $\textbf{79.63}\%$ & \cellcolor{golden!10} $\textbf{74.64}\%$\\

\hline\noalign{\smallskip}\hline\noalign{\smallskip}

\end{tabular}
}

\label{table:white_box_attack}
\end{center}
\end{table}

Throughout this work, we followed the common practice \cite{laidlaw2020perceptual, hill2020stochastic, yoon2021adversarial}, and evaluated the performance with an attack that is not exposed to the whole defense, but to the classifier alone.
We perform such an attack since attacking such an iterative defense, with hundreds of steps, is computationally challenging.
Yet, there is no empirical evidence that these defenses remain strong when attacking them with a white-box attack.

To this end, we evaluate our method with a white-box attack in \cref{table:white_box_attack}, evaluating the robustness of \AlgoNameTop under PGD\footnote{We follow the implementation of https://github.com/MadryLab/robustness} \cite{madry2017towards}. 
We use the AT method Rebuffi \emph{et al.} \cite{rebuffi2021fixing} and compare the `Base' defense to \AlgoNameTopNoSpace.
As presented, our method improves the base model, by $9\%$ and up to $29\%$ for seen and unseen attacks respectively.

We use the PGD \cite{madry2017towards} attack that operates through $20$ update steps, and we use four threat models, specified In Tab. 3. 
We evaluate the `Base' model, which is the AT model without additional test-time defense, and our method \AlgoNameTopNoSpace.

Applying a white-box attack to such an iterative defense is computationally challenging, as we need to keep in memory all of the transformation steps for all of the target classes. 
This leads to a memory consumption that grows linearly with the number of transformation steps $M$ and the number of target classes $N$. 
To overcome this difficulty, even at the cost of decreasing the robust accuracy, we use \AlgoNameTop with $k=5$ and $M=20$ transformation steps. 
Additionally, we set the hyper-parameters $\gamma=300$ and $\alpha=0.3$.

\clearpage

\section{Top-$k$ Attack}
\label{app:exp_top_k}

\AlgoNameTop is our efficient method that selects the Top-$k$ predictions from the classifier. This approach speeds up processing but might be vulnerable to adaptive attacks that exploit this specific mechanism, such as the Top-$k$ attack aimed at excluding the true class from the classifier’s Top-$k$ predictions. Typical attacks like PGD \cite{madry2017towards} focus on reducing the probability of the correct class as much as possible, which can sometimes result in outcomes similar to a Top-$k$ attack. However, achieving this is not their explicit design.

To address this, we introduce two additional adaptive attacks specifically designed to exploit the Top-$k$ vulnerability. The first, Top PGD Out (TPO), performs targeted PGD by iteratively sampling a class ranked outside of the Top-$k$ predictions and attacking towards it. The second, Top PGD In Out (TPIO), extends TPO by also removing the probability of the correct class entirely from the model’s predictions, further intensifying the attack’s focus on reducing Top-$k$ accuracy. In addition to these attacks, the RCE attack, proposed by Zhang \emph{et al.} \cite{zhang2022investigating}, is specifically designed to remove the correct class from the Top-$k$ predictions by using normalized cross-entropy loss to update logits in the direction of maximizing the rank distance. This makes RCE a suitable baseline for our evaluation alongside TPO and TPIO.

In \cref{table:top_k}, we compare the RCE attack, TPO, and TPIO to the PGD attack, specifically evaluating their effectiveness as Top-$k$ attacks by assessing the mean accuracy of the correct class appearing among the Top-$k$ predictions, rather than general model accuracy. This distinction is crucial, as the reported values reflect Top-$k$ accuracy, not the standard accuracy of the classifier in predicting the most likely class (Top-$1$ accuracy).

For example, the model by Rebuffi \cite{rebuffi2021fixing} under PGD achieves a $99.05\%$ Top-$80$ accuracy, meaning that even under a PGD attack, the true class has a $99.05\%$ likelihood of being among the top $80$ predictions of the classifier. Our results demonstrate that for k=1, the PGD attack more effectively reduces Top-$1$ accuracy, achieving a lower Top-$1$ accuracy compared to the RCE, TPO, and TPIO attacks. For higher k values, RCE perform better, evidenced by slightly lower Top-$k$ accuracy.

We want to emphasize that, despite using standard attacks for all experiments involving \AlgoNameTopNoSpace, rather than specialized Top-$k$ attacks, our results are consistently reliable. We acknowledge the potential risk that a standard attack may not fully exploit the Top-$k$ vulnerability. However, our findings clearly demonstrate that even adaptive attacks such as RCE, TPO, and TPIO, specifically designed to target Top-$k$ vulnerabilities, fail to undermine our defense.

\begin{table*}[h!]
% \vspace{-10pt}
\begin{center}
\caption{
    \textbf{Top-k Attack Analysis} A comparison of Top-k performance under a few attacks, PGD \cite{madry2017towards}, RCE \cite{zhang2022investigating} TPO and TPIO, evaluated on CIFAR100 dataset.
    Each AT model is attacked using two attacks, and the Top-k accuracy is presented for different $k$ values.
}
\resizebox{\textwidth}{!}{%

\begin{tabular}{lllllcccccccc}
\hline\noalign{\smallskip}\hline

\rowcolor{gray!5}  &  &  &  &  & \multicolumn{8}{c}{Top-k} \\
\rowcolor{gray!5} \multirow{-2}{*}{Method} & \multirow{-2}{*}{\makecell{Trained Threat \\ Model}} & \multirow{-2}{*}{Architecture} & \multirow{-2}{*}{\makecell{Attack \\Threat-Model}} & \multirow{-2}{*}{Attack Method}  & 1 & 5 & 10 & 20 & 30 & 40 & 60 & 80 \\

\multirow{4}{*}{Rebuffi et al. \cite{rebuffi2021fixing}} & \multirow{4}{*}{$L_{\infty}, \epsilon=8/255$} & \multirow{4}{*}{WRN28-10} & \multirow{4}{*}{$L_{\infty}, \epsilon=8/255$} & PGD \cite{madry2017towards} & $36.18\%$ & $64.66\%$ & $74.23\%$ & $84.44\%$ & $89.76\%$ & $93.09\%$ & $97.40\%$ & $99.29\%$ \\
 & &  &  & RCE \cite{zhang2022investigating} & $39.12\%$ & $64.48\%$ & $73.77\%$ & $83.52\%$ & $89.02\%$ & $92.22\%$ & $96.95\%$ & $99.05\%$ \\
& &  &  & TPO & $61.71\%$ & $84.18\%$ & $90.63\%$ & $95.43\%$ & $97.22\%$ & $98.43\%$ & $99.46\%$ & $99.84\%$ \\
& &  &  & TPIO & $38.56\%$ & $65.10\%$ & $74.44\%$ & $84.20\%$ & $89.61\%$ & $92.62\%$ & $97.19\%$ & $99.07\%$ \\

\hline\noalign{\smallskip}

\multirow{4}{*}{Gowal et al. \cite{gowal2020uncovering}} & \multirow{4}{*}{$L_{\infty}, \epsilon=8/255$} & \multirow{4}{*}{WRN70-16} & \multirow{4}{*}{$L_{\infty}, \epsilon=8/255$} & PGD \cite{madry2017towards} & $40.62\%$ & $69.48\%$ & $78.05\%$ & $86.30\%$ & $91.08\%$ & $93.86\%$ & $97.30\%$ & $98.94\%$  \\
 & &  &  & RCE \cite{zhang2022investigating} & $43.75\%$ & $68.47\%$ & $76.73\%$ & $84.95\%$ & $89.87\%$ & $92.81\%$ & $96.62\%$ & $98.62\%$ \\
& &  &  & TPO & $61.65\%$ & $84.38\%$ & $90.64\%$ & $95.50\%$ & $97.29\%$ & $98.53\%$ & $99.47\%$ & $99.85\%$ \\
& &  &  & TPIO & $43.73\%$ & $68.97\%$ & $77.47\%$ & $85.58\%$ & $90.52\%$ & $93.40\%$ & $96.99\%$ & $98.73\%$ \\

\hline\noalign{\smallskip}\hline\noalign{\smallskip}

\end{tabular}

}

\label{table:top_k}

\end{center}
\vspace{-20pt}
\end{table*}

\clearpage

\section{Experimental Details}
\label{app:hyper_parameters}

\setlength{\tabcolsep}{4pt}
\begin{table*}[h!]
\begin{center}
\caption{
\textbf{
\AlgoName Parameters} The parameters used for the main results presented in the paper.
}
\resizebox{0.8\textwidth}{!}{%

\begin{tabular}{llccccc}
\hline\noalign{\smallskip}\hline

Dataset & Method & Architecture & \makecell{Trained Threat Model} & $\alpha$ & $\gamma$ \\

\hline\noalign{\smallskip}

\multirow{5}{*}{CIFAR10} & Madry et al. \cite{madry2017towards} & RN50 & $L_{2}, \epsilon=0.5$ & 1.5 & 200\\

& Rebuffi et al. \cite{rebuffi2021fixing}  & WRN28-10 & $L_{2}, \epsilon=0.5$ & 0.5 & 400\\

& Rebuffi et al. \cite{rebuffi2021fixing}  & WRN28-10 & $L_{\infty}, \epsilon=8/255$ & 0.1 & 300\\

& Gowal et al. \cite{gowal2020uncovering} & WRN70-16 & $L_{\infty}, \epsilon=8/255$ & 0.3 & 300\\

& Vanila & WRN28-10 & - & 0.05 & 300\\

\hline\noalign{\smallskip}

\multirow{4}{*}{CIFAR100} & Rebuffi et al. \cite{rebuffi2021fixing} & WRN28-10 & $L_{\infty}, \epsilon=8/255$ & 0.1 & 300\\

& Rebuffi et al. \cite{rebuffi2021fixing} & WRN28-10 & $L_{\infty}, \epsilon=8/255$ & 0.1 & 300\\

& Gowal et al. \cite{gowal2020uncovering} & WRN70-16 & $L_{\infty}, \epsilon=8/255$ & 0.1 & 100\\

& Gowal et al. \cite{gowal2020uncovering} & WRN70-16 & $L_{\infty}, \epsilon=8/255$ & 0.1 & 100\\

\hline\noalign{\smallskip}

\multirow{6}{*}{ImageNet} & Madry et al. \cite{madry2017towards} & RN50 & $L_{2}, \epsilon=3.0$ & 6.0 & 5500\\

& Salman et al. \cite{salman2020adversarially}  & WRN50-2 & $L_{2}, \epsilon=3.0$ & 6.0 & 3000\\

& Madry et al. \cite{madry2017towards}  & RN50 & $L_{\infty}, \epsilon=4/255$ & 1.0 & 6000\\

& Salman et al. \cite{salman2020adversarially}  & WRN50-2 & $L_{\infty}, \epsilon=4/255$ & 1.0 & 3000\\

& Debenedetti et al. \cite{debenedetti2023light} & XCiT-S & $L_{\infty}, \epsilon=8/255$ & $0.5$ & $2500$ \\

& Debenedetti et al. \cite{debenedetti2023light} & XCiT-M & $L_{\infty}, \epsilon=8/255$ & $0.25$ & $3000$ \\

\hline\noalign{\smallskip}

\multirow{1}{*}{Flowers} & Debenedetti et al. \cite{debenedetti2023light} & XCiT-S & $L_{\infty}, \epsilon=8/255$ & $1.5$ & $500$\\

\hline\noalign{\smallskip}\hline\noalign{\smallskip}
\end{tabular}
}
\label{table:cifar10_param}
\end{center}
\end{table*}

\setlength{\tabcolsep}{4pt}
\begin{table}[h!]
\begin{center}
\caption{
\textbf{
\AlgoName Parameters} The parameters used for the results presented in the black-box experiments.
}
\resizebox{0.8\textwidth}{!}{%

\begin{tabular}{llccccc}
\hline\noalign{\smallskip}\hline

Dataset & Method & Architecture & \makecell{Trained Threat Model} & $\alpha$ & $\gamma$ \\

\hline\noalign{\smallskip}

\multirow{2}{*}{Imagenet} & Salman et al. \cite{salman2020adversarially}  & WRN50-2 & $L_{2}, \epsilon=3.0$ & 3.0 & 10000\\
 & Salman et al. \cite{salman2020adversarially}  & WRN50-2 & $L_{\infty}, \epsilon=4/255$ & 0.1 & 15000 \\

\hline\noalign{\smallskip}\hline\noalign{\smallskip}
\end{tabular}
}
% \vspace{-20pt}
\label{table:black_box_param}
\end{center}
\end{table}

\subsection{Randomized Smoothing} 
Randomized Smoothing \cite{cohen2019certified} requires that the robust classifier be trained with Gaussian noise augmentations to be effective at test-time. 
This requirement is documented in \cite{cohen2019certified} and demonstrated by CIFAR-10 results using AT \cite{madry2017towards} RN50. 
Despite its limitations, we chose to incorporate this method by selecting a small random noise value, $\sigma=0.05$, as larger values significantly degrade clean accuracy.
Additionally, we attacked this model using AutoAttack \cite{croce2020reliable} (random version), applying $10$ augmentations per image to adhere to memory constraints.
\clearpage

\section{Explain Top-$k$}
\label{app:explain_top_k}

\begin{figure}[h!]
  \centering
  \begin{subfigure}[b]{0.2\linewidth}
    \centering
    \includegraphics[width=\linewidth]{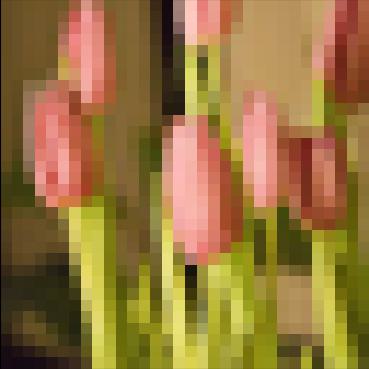}
  \end{subfigure}
  \hspace{0.02\linewidth} % Small space between images
  \begin{subfigure}[b]{0.2\linewidth}
    \centering
    \includegraphics[width=\linewidth]{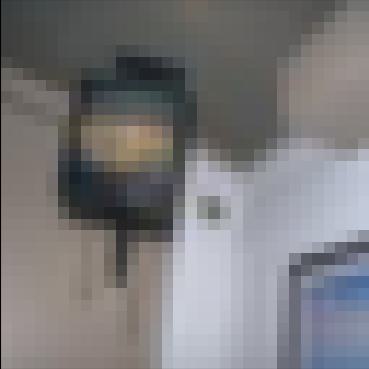}
  \end{subfigure}
  \caption{\textbf{\AlgoNameTop Success vs. CODIP Failure.} Two examples from the CIFAR-100 dataset where the Top-$k$ approach succeeds while CODIP fails. The left image shows a correctly classified tulip, while the right image demonstrates another example of correct classification by the Top-$k$ method.}
  \label{fig:explain_top_k_example}
\end{figure}

The Top-$k$ version of CODIP sometimes outperforms the full CODIP by focusing on the most likely classes, thereby reducing the influence of less relevant or noisy class transformations. By concentrating on a smaller set of high-confidence predictions, the Top-$k$ version enhances accuracy by avoiding potential noise from less likely classes. For instance, in \cref{fig:explain_top_k_example}, we show an image of a tulip (from CIFAR-100). CODIP initially predicts Oak tree followed by Tulip. However, by applying the Top-$k$ filter, Oak tree is excluded, allowing Tulip to be correctly selected. Similarly, for a television image, CODIP might predict Palm tree followed by Television. The Top-$k$ filter excludes Palm tree, leading to the accurate prediction of Television. These examples demonstrate how the Top-$k$ version of CODIP can improve performance by concentrating on the most likely classes and filtering out irrelevant ones.

\clearpage

\section{Alpha-Controlled Tradeoff for Robustness and Accuracy}
\label{app:alpha_tradeoff}

\begin{figure}[h!]
    
  \begin{center}
    \includegraphics[width=0.5\textwidth]{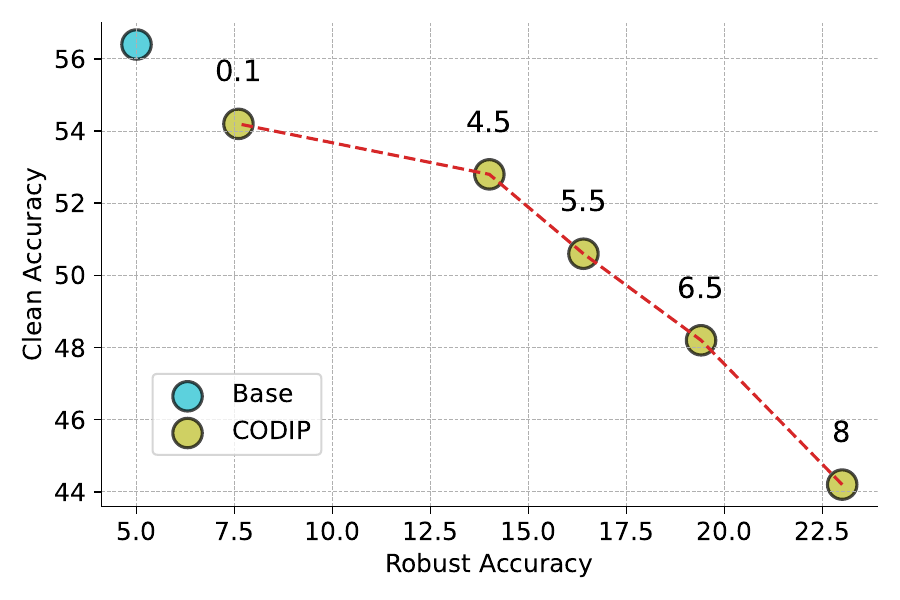}
  \end{center}
  \caption{
    \textbf{Imagenet Clean-Robust Accuracy Trade-off} A demonstration of our proposed controlled clean-robust accuracy tradeoff. 
    The tradeoff is controlled by adapting the step size value $\alpha$, specified beside each of \AlgoName workpoints.
    The used test-time methods `Base' which is the base AT model.
}
  \label{fig:app_alpha_tradeoff}
\end{figure}

The trade-off between clean and robust accuracy is controlled by the step size $\alpha$, as demonstrated not only on CIFAR-10 but also on larger datasets like ImageNet. 
In \cref{fig:app_alpha_tradeoff}, we illustrate that adjusting $\alpha$ allows for flexible control over this trade-off on ImageNet, further validating our approach. 
Additionally, while Tab. 1 and Tab. 2 show that clean accuracy may drop, this drop is manageable and adjustable by fine-tuning $\alpha$. 
This supports our claim that CODIP can effectively manage the clean-robust accuracy balance, even on more complex datasets with more classes, such as ImageNet, by adjusting the value of $\alpha$.

Note that \cref{fig:app_alpha_tradeoff} was evaluated on a random subset of $500$ examples from the ImageNet validation set, as it is intended for exemplification purposes.

\clearpage

\section{$\gamma$ Effect on Clean-Robust Accuracy Trade-off}
\label{app:gamma_tradeoff}

As demonstrated in Fig. 3, the hyperparameter $\gamma$ plays a critical role in controlling the clean-robust accuracy trade-off across different values of $\alpha$. Both $\alpha$ and $\gamma$ contribute to restricting the transformation in order to maintain the accuracy of the model. While $\alpha$ primarily governs the magnitude of the perturbation, $\gamma$ regularizes the transformation by ensuring that it remains subtle enough to preserve the natural characteristics of the image while successfully changing its class, even in the presence of small-norm perturbations.

\begin{figure*}[h!]
  \begin{center}
  \vspace{-12pt}
    \includegraphics[width=0.4\textwidth]{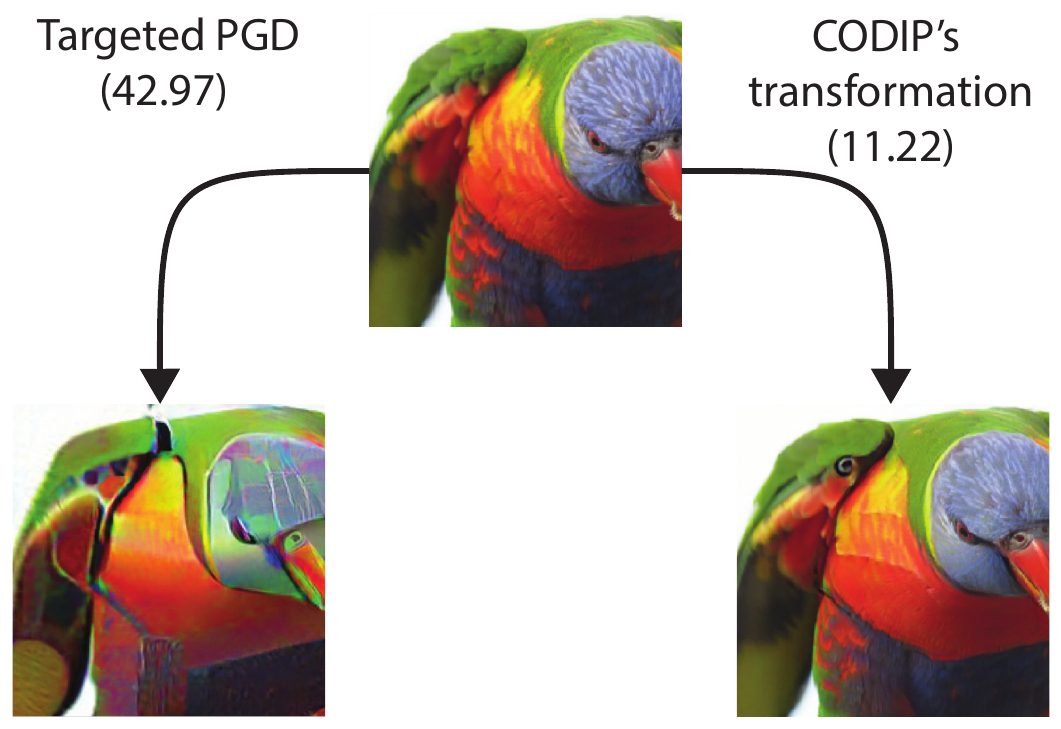}
  \end{center}
  \vspace{-10pt}
  \caption{
    \textbf{\AlgoName vs. Targeted PGD} A comparison between \AlgoName and targeted PGD class conditioned transformations. 
    An image of \emph{lorikeet} is transformed into a \emph{toucan} using \AlgoName and targeted PGD. A $\ell_2$ distance between the clean image and the transformed ones is stated beneath the attack name.
}
  \label{fig:CODIP_vs_PGD}
\end{figure*}

When $\gamma = 0$, the transformation is unregularized, leading to a situation where both clean and robust accuracy suffer significantly. Without the restriction provided by $\gamma$, the transformation can deviate excessively from the input image, negatively impacting both clean accuracy and robustness. This issue is illustrated in \cref{fig:CODIP_vs_PGD}, where we compare transformations produced by \AlgoName and targeted PGD. The image transformed by \AlgoName (right) maintains a closer resemblance to the original input image, while the image generated by targeted PGD (left) appears unnatural and distorted. The $\ell_2$ distances further highlight this: \AlgoName achieves a much lower distance (11.22) compared to targeted PGD (42.97). In addition, in Fig. 3, we show that both robust and clean accuracy metrics are substantially worse when $\gamma = 0$, emphasizing that an unregularized transformation fails to achieve the desired balance between robustness and preserving the integrity of the original input.

To achieve the optimal balance between clean and robust accuracy, it is essential to carefully tune $\gamma$. In Fig. 3, we present three graphs that illustrate the impact of different $\gamma$ values across various $\alpha$ settings. In each experiment, increasing $\gamma$ initially improves clean accuracy. For larger $\alpha$ values, increasing $\gamma$ significantly enhances robustness, though after a certain threshold, clean accuracy may begin to slightly decrease. For smaller $\alpha$, robust accuracy increases slightly before eventually decreasing. These findings demonstrate that while careful tuning of $\gamma$ is necessary, it remains a critical parameter for managing the clean-robust trade-off, as it controls how closely the transformation adheres to the original image, especially under varying perturbation magnitudes. Specifically, we show its pivotal role in controlling clean accuracy

% As demonstrated in Fig. 3, the hyperparameter $\gamma$ plays a crucial role in regulating the clean-robust accuracy trade-off across different values of $\alpha$. While both $\alpha$ and $\gamma$ seem to serve a similar purpose—restricting the transformation to maintain accuracy—$\gamma$ proves essential even when $\alpha$ is small. Specifically, for small $\alpha$, increasing $\gamma$ significantly improves clean accuracy without substantially affecting robustness. This is because the conditioned transformation remains necessary even for small-norm transformations to prevent excessive changes to the input.

% For larger $\alpha$, $\gamma$ enhances both robustness and clean accuracy initially, but eventually, over-regularization leads to a slight decline in clean accuracy. This behavior shows that $\gamma$ must be carefully tuned, even when $\alpha$ is small, to ensure the transformation stays effective without compromising the clean-robust trade-off. Thus, $\gamma$ is key for modulating the transformation across various $\alpha$ values, particularly in maintaining clean accuracy in scenarios where small-norm transformations are applied.

% In the next experiment, we emphasize the transformation benefits of \AlgoName, as we compare our method to targeted PGD on \cref{fig:CODIP_vs_PGD}.
% Both targeted PGD and \AlgoName change the image's appearance, trying to change its predicted class. 
% However, targeted PGD  creates an image that does not look natural,
% while \AlgoName creates a natural-looking image, as it keeps the transformed image pixel-wise close to the input image.

\clearpage

%%%%%%%%% REFERENCES
{\small
\bibliographystyle{ieee_fullname}
\bibliography{egbib}
}